\documentclass[10pt,twocolumn,letterpaper]{article}

\usepackage{iccv}
\usepackage{times}
\usepackage{epsfig}
\usepackage{graphicx}
\usepackage{amsmath}
\usepackage{amssymb}
\usepackage{xcolor}
\usepackage{pifont}

\usepackage[accsupp]{axessibility}
\usepackage[pagebackref=true,breaklinks=true,colorlinks=true,bookmarks=false]{hyperref}
\usepackage[capitalize,noabbrev]{cleveref}

\usepackage{colortbl}
\usepackage{booktabs, adjustbox}
\usepackage{xcolor} %
\usepackage{multirow}
\usepackage{enumitem}
\usepackage{algorithm}
\RequirePackage{algorithmic}

\newcommand{\model}{Diffusion Classifier\xspace}

\newcommand\blfootnote[1]{%
  \begingroup
  \renewcommand\thefootnote{}\footnote{#1}%
  \addtocounter{footnote}{-1}%
  \endgroup
}

\newcommand{\xmark}{\textcolor{red}{\ding{55}}}
\newcommand{\greencheck}{\textcolor{green}{$\checkmark$}}
\DeclareMathOperator*{\argmax}{arg\,max}
\DeclareMathOperator*{\argmin}{arg\,min}

\definecolor{first}{rgb}{1.0, 0.93, 0.7}

\definecolor{dt}{gray}{0.7}  
\def \first {\cellcolor{green!15}}

\makeatletter
\def\blfootnote{\gdef\@thefnmark{}\@footnotetext}
\makeatother

\ifx \cvprsubmission \undefined
    \newcommand{\todo}[1]{\textcolor{red}{todo: #1}}

\else
	\newcommand{\todo}[1]{}

\fi

\iccvfinalcopy %

\def\iccvPaperID{11420} %

\begin{document}

\title{Your Diffusion Model is Secretly a Zero-Shot Classifier}

\author{Alexander C. Li$\quad$
Mihir Prabhudesai$\quad$
Shivam Duggal$\quad$
Ellis Brown$\quad$
Deepak Pathak
\vspace{2.5mm} \\
Carnegie Mellon University
}
\maketitle
\blfootnote{Correspondence to: Alexander Li $<$\href{mailto:alexanderli@cmu.edu}{alexanderli@cmu.edu}$>$} 

\begin{abstract}
The recent wave of large-scale text-to-image diffusion models has dramatically increased our text-based image generation abilities.
These models can generate realistic images for a staggering variety of prompts and exhibit impressive compositional generalization abilities. 
Almost all use cases thus far have solely focused on sampling; however, diffusion models can also provide conditional density estimates, which are useful for tasks beyond image generation.
In this paper, we show that the density estimates from large-scale text-to-image diffusion models like Stable Diffusion can be leveraged to perform zero-shot classification \textbf{without any additional training}.
Our generative approach to classification, which we call \textbf{Diffusion Classifier},
attains strong results on a variety of benchmarks and outperforms alternative methods of extracting knowledge from diffusion models. 
Although a gap remains between generative and discriminative approaches on zero-shot recognition tasks, our diffusion-based approach has significantly stronger multimodal compositional reasoning ability than competing discriminative approaches. 
Finally, we use Diffusion Classifier to extract standard classifiers from class-conditional diffusion models trained on ImageNet. Our models achieve strong classification performance using only weak augmentations and exhibit qualitatively better ``effective robustness'' to distribution shift. Overall, our results are a step toward using generative over discriminative models for downstream tasks.
Results and visualizations on our website: \href{https://diffusion-classifier.github.io/}{\url{diffusion-classifier.github.io/}}
\end{abstract}

\vspace{-0.1in}
\section{Introduction}
\textit{To Recognize Shapes, First Learn to Generate Images}~\cite{Hinton2007ToRS}---in this seminal paper, Geoffrey Hinton emphasizes generative modeling as a crucial strategy for training artificial neural networks for discriminative tasks like image recognition.  Although generative models tackle the more challenging task of accurately modeling the underlying data distribution, they can create a more complete representation of the world that can be utilized for various downstream tasks. As a result, a plethora of implicit and explicit generative modeling approaches \cite{goodfellow2014generative,KingmaWelling,ebm_lecun,realNVP,PixelRNN, sohl2015deep, VincetScoreMatching} have been proposed over the last decade. However, the primary focus of these works has been content creation \cite{devlin2018bert, brown2020language, pix2pix2017, StyleGAN, wavenet, imagen_video} rather than their ability to perform discriminative tasks.
In this paper, we revisit this classic generative vs.\ discriminative debate in the context of diffusion models, the current state-of-the-art generative model family.
In particular, we examine \textit{how diffusion models compare against the state-of-the-art discriminative models on the task of image classification.}

Diffusion models are a recent class of likelihood-based generative models that model the data distribution via an iterative noising and denoising procedure~\cite{sohl2015deep, ho2020denoising}.
They have recently achieved state-of-the-art performance \cite{dhariwal2021diffusion} on several text-based content creation and editing tasks \cite{ramesh2022, saharia2022photorealistic,imagen_video, ruiz2022dreambooth, poole2022dreamfusion}.  Diffusion models operate by performing two iterative processes---the fixed \textit{forward process}, which destroys structure in the data by iteratively adding noise, and the learned \textit{backward process}, which attempts to recover the structure in the noised data. These models are trained via a variational objective, which maximizes an evidence lower bound (ELBO)~\cite{variational_inference} of the log-likelihood. 
For most diffusion models, computing the ELBO consists of adding noise $\epsilon$ to a sample, using the neural network to predict the added noise, and measuring the prediction error.

\begin{figure*}[t]
    \centering
    \includegraphics[width=\linewidth]{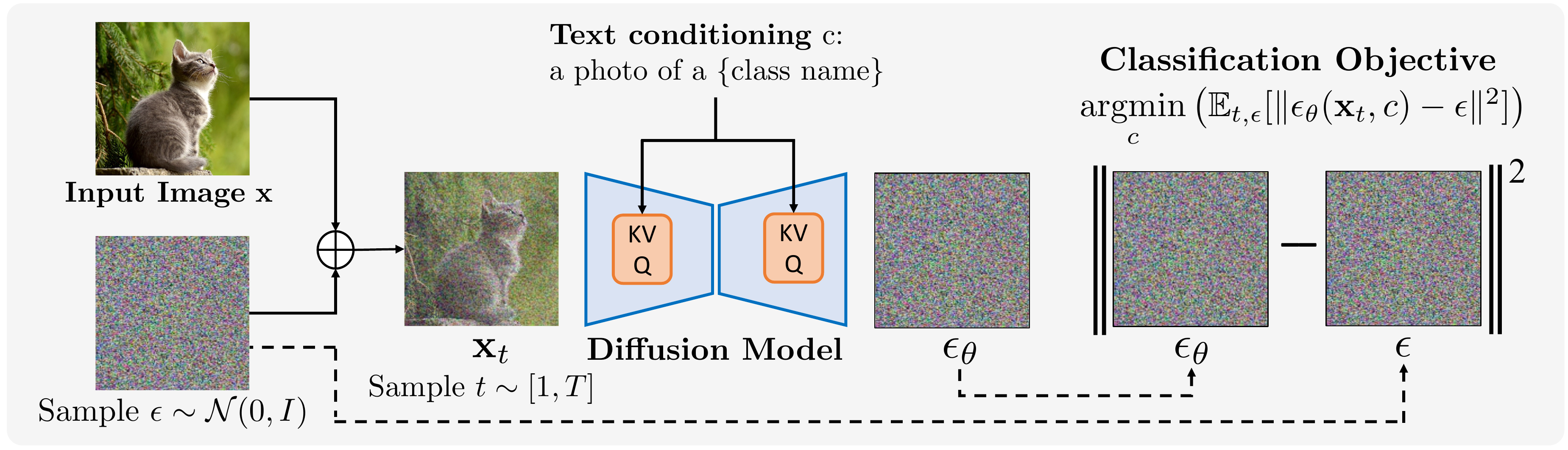}
    \vspace{-3.5mm}
    \caption{\textbf{Overview of our Diffusion Classifier approach:} Given an input image $\mathbf{x}$ and a set of possible conditioning inputs (e.g., text for Stable Diffusion or class index for DiT, an ImageNet class-conditional model), we use a diffusion model to choose the one that best fits this image. \model is theoretically motivated through the variational view of diffusion models and uses the ELBO to approximate $\log p_\theta(\mathbf{x}\mid \mathbf{c})$. \model chooses the conditioning $\mathbf{c}$ that best predicts the noise added to the input image. \textit{\model can be used to extract a zero-shot classifier from Stable Diffusion and a standard classifier from DiT without any additional training.}}
    \label{fig:method}
    \vspace{-2mm}
\end{figure*}

Conditional generative models like diffusion models can be easily converted into classifiers~\cite{NgJordon}. Given an input $\mathbf{x}$ and a finite set of classes $\mathbf{c}$ that we want to choose from, we can use the model to compute class-conditional likelihoods $p_\theta(\mathbf{x} \mid \mathbf{c})$. Then, by selecting an appropriate prior distribution $p(\mathbf{c})$ and applying Bayes' theorem, we can get predicted class probabilities $p(\mathbf{c} \mid \mathbf{x})$. 
For conditional diffusion models that use an auxiliary input, like a class index for class-conditioned models or prompt for text-to-image models, we can do this by leveraging the ELBO as an approximate class-conditional log-likelihood $\log p(\mathbf{x} \mid \mathbf{c})$. 
In practice, obtaining a diffusion model classifier through Bayes' theorem consists of repeatedly adding noise and computing a Monte Carlo estimate of the expected noise reconstruction losses (also called $\epsilon$-prediction loss) for every class.  
We call this approach \textbf{Diffusion Classifier}. 
Diffusion Classifier can extract zero-shot classifiers from text-to-image diffusion models and standard classifiers from class-conditional diffusion models, \textit{without any additional training}.
We develop techniques for appropriately choosing diffusion timesteps to compute errors at, reducing variance in the estimated probabilities, and speeding up classification inference.

We highlight the surprising effectiveness of our proposed Diffusion Classifier on zero-shot classification, compositional reasoning, and supervised classification tasks by comparing against multiple baselines on eleven different benchmarks. By utilizing Stable Diffusion~\cite{rombach2022high}, Diffusion Classifier achieves strong zero-shot accuracy and outperforms alternative approaches for extracting knowledge from the pretrained diffusion model. 
Our approach also \textit{outperforms the strongest contrastive methods on the challenging Winoground compositional reasoning benchmark}~\cite{thrush2022winoground}. 
Finally, we use our approach to perform standard classification with Diffusion Transformer (DiT), an ImageNet-trained class-conditional diffusion model. Our generative approach \textit{achieves 79.1\% accuracy on ImageNet using only weak augmentations and exhibits better robustness to distribution shift} than competing discriminative classifiers trained on the same dataset. Our results suggest that it may be time to revisit generative approaches to classification.

\section{Related Work}
\vspace{-0.05in}
\paragraph{Generative Models for Discriminative Tasks:} Machine learning algorithms designed to solve common classification or regression tasks generally operate under two paradigms: \textit{discriminative} approaches directly learn to model the decision boundary of the underlying task, while \textit{generative approaches} learn to model the distribution of the data and then address the underlying task as a maximum likelihood estimation problem. Algorithms like naive Bayes \cite{NgJordon}, VAEs \cite{KingmaWelling}, GANs \cite{goodfellow2014generative}, EBMs \cite{Du2019ImplicitGA, ebm_lecun}, and diffusion models \cite{sohl2015deep, ho2020denoising} fall under the category of generative models. The idea of modeling the data distribution to better learn the discriminative feature has been highlighted by several seminal works \cite{Hinton2007ToRS, NgJordon, RanzatoHinton}. These works train deep belief networks \cite{deep_belief_networks} to model the underlying image data as latents, which are later used for image recognition tasks. Recent works on generative modeling have also learned efficient representations for both global and dense prediction tasks like classification \cite{MaskedAutoencoders2021, hjelm2018learning, croce-etal-2020-gan, brown2020language, Devlin2019BERTPO} and segmentation \cite{semanticGAN, zhang21, InfoGAN, baranchuk2022labelefficient,burgert2022peekaboo}. Moreover, such models \cite{grathwohl2020your, Liu2020HybridDT, NEURIPS2020_0660895c} have been shown to be more adversarially robust and better calibrated. However, most of the aforementioned works either train jointly for discriminative and generative modeling or fine-tune generative representations for downstream tasks. Directly utilizing generative models for discriminative tasks is a relatively less-studied problem, and in this work, we  particularly highlight the \textit{efficacy of directly using recent diffusion models as image classifiers.}

\vspace{-0.1in}
\paragraph{Diffusion Models:}

Diffusion models \cite{ho2020denoising, sohl2015deep} have recently gained significant attention from the research community due to their ability to generate high-fidelity and diverse content like images \cite{saharia2022photorealistic, glide, ramesh2022}, videos \cite{singer2023makeavideo, imagen_video, Villegas2022PhenakiVL}, 3D \cite{poole2022dreamfusion, lin2022magic3d}, and audio \cite{kong2021diffwave, liu2023audioldm} from various input modalities like text.
Diffusion models are also closely tied to EBMs \cite{ebm_lecun, Du2019ImplicitGA}, denoising score matching \cite{SongErmon, vincent2008extracting}, and stochastic differential equations \cite{song2020score,zimmermann2021score}.
In this work, we investigate to what extent the impressive high-fidelity generative abilities of these diffusion models can be utilized for discriminative tasks (namely classification). We take advantage of the variational view of diffusion models for efficient and parallelizable density estimates. 
The prior work of Dhariwal \& Nichol \cite{dhariwal2021diffusion} proposed using a classifier network to modify the output of an unconditional generative model to obtain class-conditional samples. Our goal is the reverse: using diffusion models as classifiers.  %

\vspace{-0.1in}
\paragraph{Zero-Shot Image Classification:}
Classifiers thus far have usually been trained in a supervised setting where the train and test sets are fixed and limited.  CLIP \cite{radford2019language} showed that exploiting large-scale image-text data can result in zero-shot generalization to various new tasks. 
Since then, there has been a surge toward building a new category of classifiers, known as zero-shot or open-vocabulary classifiers, that are capable of detecting a wide range of class categories \cite{gadre2022clip,li2022blip,li2022grounded,alayrac2022flamingo}. These methods have been shown to learn robust representations that generalize to various distribution shifts \cite{ilharco2021openclip,dehghani2023scaling,taori2020measuring}.
Note that in spite of them being called ``zero-shot,''  it is still unclear whether evaluation samples lie in their training data distribution. 
In contrast to the discriminative approaches above, we propose extracting a zero-shot classifier from a large-scale \textit{generative} model.

\section{Method: Classification via Diffusion Models}
\label{sec:method}

We describe our approach for calculating class conditional density estimates in a practical and efficient manner using diffusion models. We first provide an overview of diffusion models (Sec.~\ref{subsec:diffusion_background}), discuss the motivation and derivation of our \model method (Sec.~\ref{subsec:derivation}), and finally propose techniques to improve its accuracy (Sec.~\ref{subsec:paired_diff}).

\subsection{Diffusion Model Preliminaries}
\label{subsec:diffusion_background}
Diffusion probabilistic models (``diffusion models'' for short) \cite{sohl2015deep,ho2020denoising} are generative models with a specific Markov chain structure. Starting at a clean sample $\mathbf{x}_0$, the fixed forward process $q(\mathbf{x}_t \mid \mathbf{x}_{t-1})$  adds Gaussian noise, whereas the learned reverse process $p_\theta(\mathbf{x}_{t-1} \mid \mathbf{x}_t, \mathbf{c})$ tries to denoise its input, optionally conditioning on a variable $\mathbf{c}$. In our setting, $\mathbf{x}$ is an image and $\mathbf{c}$ represents a low-dimensional text embedding (for text-to-image synthesis) or class index (for class-conditional generation). Diffusion models define the conditional probability of $\mathbf{x}_0$ as:
\begin{align}
p_\theta(\mathbf{x}_0 \mid \mathbf{c}) = \int_{\mathbf{x}_{1:T}} p(\mathbf{x}_T) \prod_{t=1}^T p_\theta(\mathbf{x}_{t-1} \mid \mathbf{x}_t, \mathbf{c})\  \mathrm{d}\mathbf{x}_{1:T}
\end{align}
where $p(\mathbf{x}_T)$ is typically fixed to $\mathcal{N}(0, I)$. Directly maximizing $p_\theta(\mathbf{x}_0)$ is intractable due to the integral, so diffusion models are instead trained to minimize the variational lower bound (ELBO) of the log-likelihood:
\begin{align}
   \log p_\theta (\mathbf{x}_0 \mid \mathbf{c}) & \geq \mathbb{E}_{q}\left[\log \frac{p_\theta(\mathbf{x}_{0:T}, \mathbf{c})}{q(\mathbf{x}_{1:T} \mid \mathbf{x}_0)}\right]
\end{align}
Diffusion models parameterize $p_\theta(\mathbf{x}_{t-1}\mid \mathbf{x}_t, \mathbf{c})$ as a Gaussian and train a neural network to map a noisy input $\mathbf{x}_t$ to a value used to compute the mean of $p_\theta(\mathbf{x}_{t-1}\mid \mathbf{x}_t, \mathbf{c})$. Using the fact that 
each noised sample
$\mathbf{x}_t =\sqrt{\bar \alpha_{t}}\mathbf{x} + \sqrt{1-\bar\alpha_{t}} \epsilon$ can be written as a weighted combination of a clean input $\mathbf{x}$ and Gaussian noise $\epsilon \sim \mathcal{N}(0, I)$, diffusion models typically learn a network $\epsilon_\theta(\mathbf{x}_t, \mathbf{c})$ that estimates the added noise. Using this parameterization, the ELBO can be written as:
\begin{align}
   - \mathbb{E}_\epsilon \left[\sum_{t = 2}^T w_t \|\epsilon - \epsilon_\theta(\mathbf{x}_t, \mathbf{c})\|^2 - \log p_\theta(\mathbf{x}_0 \mid \mathbf{x}_1, \mathbf{c}) \right] + C
\label{eq:exact_elbo}
\end{align}
where $C$ is a constant term that does not depend on $\mathbf{c}$. Since $T = 1000$ is large and $\log p_\theta(\mathbf{x}_0 \mid \mathbf{x}_1, \mathbf{c})$ is typically small, we choose to drop this term. Finally, \cite{ho2020denoising} find that removing $w_t$ improves sample quality metrics, and many follow-up works also choose to do so. We found that deviating from the uniform weighting used at training time hurts accuracy, so we set $w_t = 1$.
Thus, this gives us our final approximation that we treat as the ELBO: 
\begin{align}
    - \mathbb{E}_{t, \epsilon} \left[\|\epsilon - \epsilon_\theta(\mathbf{x}_t, \mathbf{c})\|^2 \right] + C
\label{eq:elbo}
\end{align}

\renewcommand{\algorithmiccomment}[1]{// #1}

\begin{algorithm}[t]
   \caption{$\texttt{Diffusion Classifier}$}
\label{alg:diffusion_classifier_naive}
\begin{algorithmic}[1]
    \STATE {\bfseries Input:}
    test image $\mathbf{x}$,
    conditioning inputs 
    $\{\mathbf{c}_i\}_{i=1}^n$
    (\eg, text embeddings),
    \# of trials $T$ per input
    \STATE Initialize $\texttt{Errors}[\mathbf{c}_i] = \text{list}()$ for each $\mathbf{c}_i$ 
    \FOR{trial $j = 1, \dots, T$}
        \STATE Sample $t \sim [1, 1000]$; \  $\epsilon \sim \mathcal{N}(0, I)$
        \STATE $\mathbf{x}_t = \sqrt{\bar \alpha_{t}}\mathbf{x} + \sqrt{1-\bar\alpha_{t}} \epsilon$
        \FOR{conditioning $\mathbf{c}_k \in \{\mathbf{c}_i\}_{i=1}^n$}
            \STATE $\texttt{Errors}[\mathbf{c}_k].\texttt{append}(\|\epsilon - \epsilon_\theta(\mathbf{x}_t, \mathbf{c}_k)\|^2)$
        \ENDFOR
    \ENDFOR
    \RETURN $\displaystyle \argmin_{\mathbf{c}_i \in \mathcal C} \texttt{\,mean}(\texttt{Errors}[\mathbf{c}_i])$
\end{algorithmic}
\end{algorithm}

\begin{figure}[t]
    \centering
    \includegraphics[width=\linewidth]{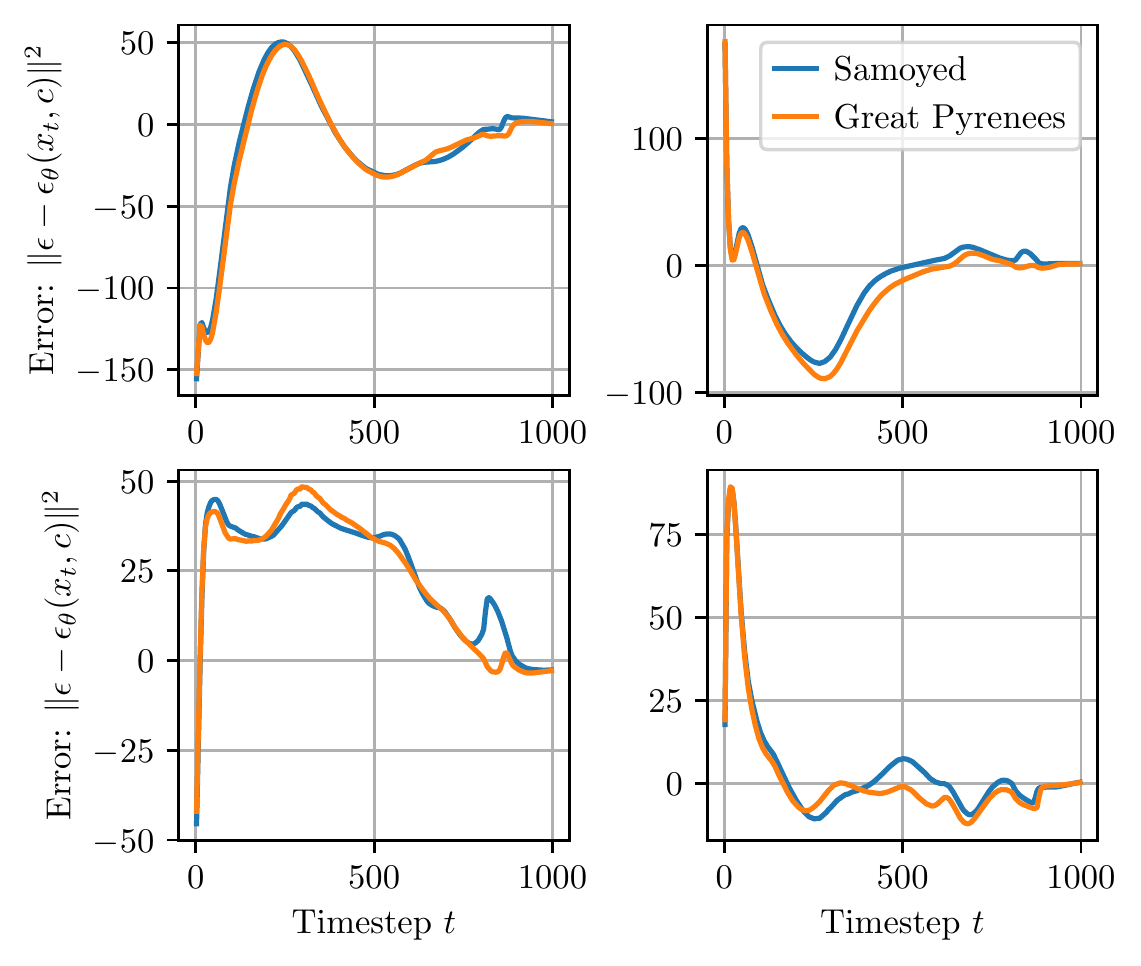}
    \vspace{-0.2in}
    \caption{We show the $\epsilon$-prediction error for an image of a Great Pyrenees dog and two prompts (``Samoyed'' and ``Great Pyrenees''). Each subplot corresponds to a single $\epsilon_i$, with the error evaluated at every $ t \in \{1, 2, ..., 1000\}$. Errors are normalized to be zero-mean at each timestep across the 4 plots, and lower is better. Variance in $\epsilon$-prediction error is high across different $\epsilon$, but the variance in the error difference between prompts is much smaller.}
    \label{fig:correlated_errors}
    \vspace{-0.025in}
\end{figure}

\subsection{Classification with diffusion models}
\label{subsec:derivation}
In general, classification using a conditional generative model can be done by using Bayes' theorem on the model predictions $p_\theta(\mathbf{x} \mid \mathbf{c}_i)$ and the prior $p(\mathbf{c})$ over labels $\{\mathbf{c}_i\}$: 
\begin{align}
    p_\theta(\mathbf{c}_i \mid \mathbf{x}) = \frac{p(\mathbf{c}_i)\ p_\theta(\mathbf{x} \mid \mathbf{c}_i)}{\sum_j p(\mathbf{c}_j)\ p_\theta(\mathbf{x} \mid \mathbf{c}_j)}
\label{eq:bayes}
\end{align}
A uniform prior over $\{\mathbf{c}_i\}$ (i.e., $p(\mathbf{c}_i) = \frac{1}{N}$) is natural and leads to all of the $p(\mathbf{c})$ terms cancelling. For diffusion models, computing $p_\theta(\mathbf{x}\mid \mathbf{c})$ is intractable, so we use the ELBO in place of $\log p_\theta(\mathbf{x} \mid \mathbf{c})$ and use Eq.~\ref{eq:elbo} and Eq.~\ref{eq:bayes} to obtain a posterior distribution over $\{\mathbf{c}_i\}_{i=1}^N$:
\begin{align}
    p_\theta(\mathbf{c}_i \mid  \mathbf{x}) 
    &= \frac{\exp\{- \mathbb{E}_{t, \epsilon}[\|\epsilon - \epsilon_\theta(\mathbf{x}_t, \mathbf{c}_i)\|^2]\}}{\sum_j \exp\{- \mathbb{E}_{t, \epsilon}[\|\epsilon - \epsilon_\theta(\mathbf{x}_t, \mathbf{c}_j)\|^2]\}} 
\label{eq:posterior}
\end{align}
We compute an unbiased Monte Carlo estimate of each expectation by sampling $N$ $(t_i, \epsilon_i)$ pairs, with $t_i \sim [1, 1000]$ and $\epsilon \sim \mathcal{N}(0, I)$, and computing:
\begin{align}
    \frac{1}{N}\sum_{i=1}^N \left\|\epsilon_i - \epsilon_\theta(\sqrt{\bar \alpha_{t_i}}\mathbf{x} + \sqrt{1-\bar\alpha_{t_i}} \epsilon_i, \mathbf{c}_j)\right\|^2
\label{eq:monte_carlo}
\end{align}
By plugging Eq.~\ref{eq:monte_carlo} into Eq.~\ref{eq:posterior}, we can extract a classifier from any conditional diffusion model. We call this method \textbf{\model}.
\textit{\model is a powerful, hyperparameter-free approach to extracting classifiers from pretrained diffusion models without any additional training.} 
\model can be used to extract a zero-shot classifier from a text-to-image model like Stable Diffusion \cite{rombach2022high}, to extract a standard classifier from a class-conditional diffusion model like DiT \cite{Peebles2022DiT}, and so on. 
We outline our method in \cref{alg:diffusion_classifier_naive} and show an overview in \cref{fig:method}.

\subsection{Variance Reduction via Difference Testing}
\label{subsec:paired_diff}
At first glance, it seems that accurately estimating $\mathbb{E}_{t, \epsilon}\left[\|\epsilon - \epsilon_\theta(\mathbf{x}_t, \mathbf{c})\|^2 \right]$ for each class $\mathbf{c}$ requires prohibitively many samples. Indeed, a Monte Carlo estimate even using thousands of samples is not precise enough to distinguish classes reliably. However, a key observation is that classification only requires the \textit{relative} differences between the prediction errors, not their \textit{absolute} magnitudes. We can rewrite the approximate $p_\theta(\mathbf{c}_i \mid \mathbf{x})$ from Eq.~\ref{eq:posterior} as:
\begin{align}
    \frac{1}{\sum_j \exp\left\{\mathbb{E}_{t, \epsilon}[\|\epsilon -\epsilon_\theta(\mathbf{x}_t, \mathbf{c}_i)\|^2 -\|\epsilon -\epsilon_\theta(\mathbf{x}_t, \mathbf{c}_j)\|^2] \right\}}
    \label{eq:paired}
\end{align}
Eq.~\ref{eq:paired} makes apparent that we only need to estimate the \emph{difference} in prediction errors across each conditioning value. Practically, instead of using different random samples of  $(t_i, \epsilon_i)$ to estimate the ELBO for each conditioning input $\mathbf{c}$, we simply sample a fixed set $S = \{(t_i, \epsilon_i)\}_{i=1}^N$ and use the same samples to estimate the $\epsilon$-prediction error for every $\mathbf{c}$.
This is reminiscent of paired difference tests in statistics, which increase their statistical power by matching conditions across groups and computing differences.

In Figure~\ref{fig:correlated_errors}, we use 4 fixed $\epsilon_i$'s and evaluate $\|\epsilon_i - \epsilon_\theta(\sqrt{\bar \alpha_{t}}\mathbf{x} + \sqrt{1-\bar\alpha_{t}} \epsilon_i, \mathbf{c}) \|^2$ for every $t \in {1, \dots, 1000}$, two prompts (``Samoyed dog'' and ``Great Pyrenees dog''), and a fixed input image of a Great Pyrenees. Even for a fixed prompt, the $\epsilon$-prediction error varies wildly across the specific $\epsilon_i$ used. However, the error difference between each prompt is much more consistent for each $\epsilon_i$. \textit{Thus, by using the same $(t_i, \epsilon_i)$ for each conditioning input, our estimate of $p_\theta(\mathbf{c}_i\mid \mathbf{x})$ is much more accurate. }

\begin{figure}
    \centering
    \includegraphics[width=\linewidth]{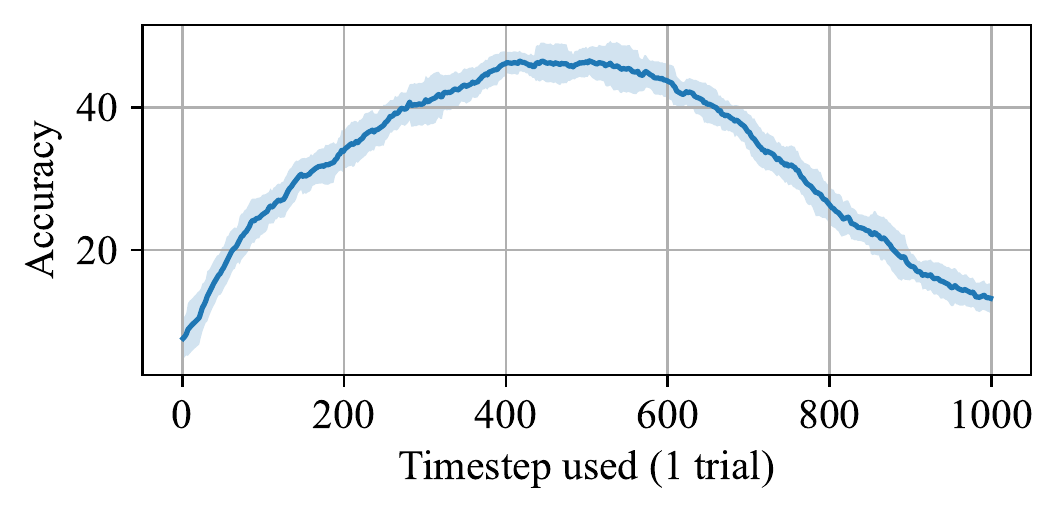}
    \vspace{-0.2in}
    \caption{\textbf{Pets accuracy, evaluating only a single timestep per class}. Small $t$ corresponds to less noise added, and large $t$ corresponds to significant noise. Accuracy is highest when an intermediate amount of noise is added ($t=500$).}
    \label{fig:single_t}
\end{figure}

\section{Practical Considerations}
\label{sec:practical}
Our \model method requires repeated error prediction evaluations for every class in order to classify an input image. These evaluations naively require significant inference time, even with the technique presented in \cref{subsec:paired_diff}. In this section, we present further insights and optimizations that reduce our method's runtime.

\subsection{Effect of timestep}

Diffusion Classifier, which is a theoretically principled method for estimating $p_\theta(\mathbf{c}_i \mid  \mathbf{x})$, uses a uniform distribution over the timestep $t$ for estimating the $\epsilon$-prediction error. Here, we check if alternate distributions over $t$ yield more accurate results. Figure~\ref{fig:single_t} shows the Pets accuracy when using only a single timestep evaluation per class. Perhaps intuitively, accuracy is highest when using intermediate timesteps ($t \approx 500)$. This begs the question: can we improve accuracy by oversampling intermediate timesteps and undersampling low or high timesteps?

\begin{figure}[t]
    \centering
    \includegraphics[width=0.9\linewidth]{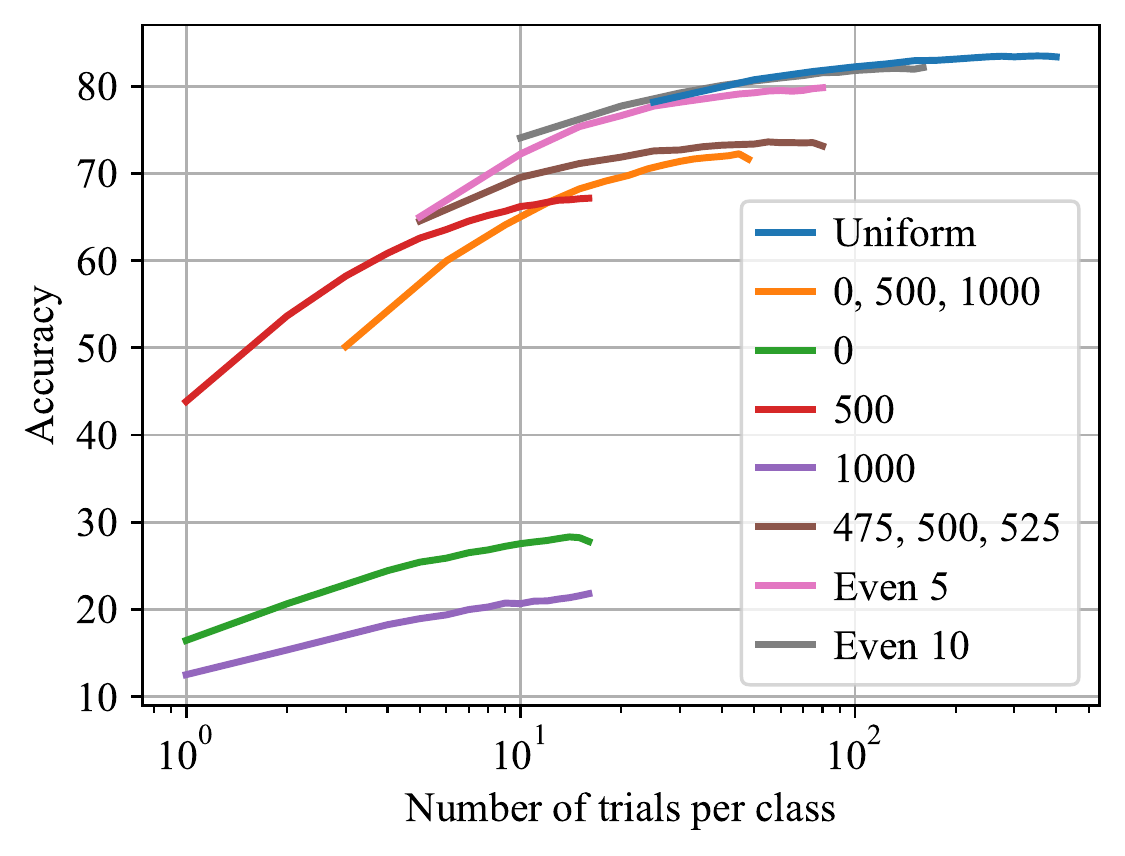}
    \caption{\small \textbf{Zero-shot scaling curves for different timestep sampling strategies}. We evaluate a variety of strategies for choosing the timesteps at which we evaluate the $\epsilon$-prediction error. Each strategy name indicates which timesteps it uses---
    e.g., ``$0$'' only uses the first timestep, ``$0,500,1000$'' uses only the first, middle and last, ``Even 10'' uses 10 evenly spaced timesteps.
    We allocate more $\epsilon$ evaluations at the chosen timesteps as the number of trials increases. Strategies that repeatedly sample from a restricted set of timesteps, like ``475, 500, 525'', scale poorly with trials. Using timesteps uniformly from the full range [1, 1000] scales best.
    }
    \label{fig:scaling}
    \vspace{-2mm}
\end{figure}

We try a variety of timestep sampling strategies, including repeatedly trying $t=500$ with many random $\epsilon$, trying $N$ evenly spaced timesteps, and trying the middle ${t - N/2, \dots, t + N/2}$ timesteps. The tradeoff between different strategies is whether to try a few $t_i$ repeatedly with many $\epsilon$ or to try many $t_i$ once. Figure~\ref{fig:scaling} shows that all strategies improve when taking using average error of more samples, but simply using evenly spaced timesteps is best. 
We hypothesize that repeatedly trying a small set of $t_i$ scales poorly since this biases the ELBO estimate.

\subsection{Efficient Classification}
\label{sec:efficient_classification}
A naive implementation of our method requires $C \times N$ trials to classify a given image, where $C$ is the number of classes and $N$ is the number of $(t, \epsilon)$ samples to evaluate to estimate each conditional ELBO. However, we can do better. Since we only care about $\argmax_{\mathbf{c}} p(\mathbf{c}\mid \mathbf{x})$, we can stop computing the ELBO for classes we can confidently reject. Thus, one option to classify an image is to use an upper confidence bound algorithm~\cite{auer2002using} to allocate most of the compute to the top candidates. However, this requires assuming that the distribution of $\|\epsilon -\epsilon_\theta(\mathbf{x}_t, \mathbf{c}_j)\|^2$ is the same across timesteps $t$, which does not hold. 

We found that a simpler method works just as well.
We split our evaluation into a series of stages, where in each stage we try each remaining $\mathbf{c}_i$ some number of times and then remove the ones that have the highest average error.
This allows us to efficiently eliminate classes that are almost certainly not the final output and allocate more compute to reasonable classes. For example, on the Pets dataset, we have $N_{\text{stages}}$ = 2. We try each class 25 times in the first stage, then prune to the 5 classes with the smallest average error. Finally, in the second stage we try each of the 5 remaining classes 225 additional times.
In \cref{alg:diffusion_classifier_accel}, we write this as $\texttt{KeepList} = (5, 1)$ and $\texttt{TrialList} = (25, 250)$.
With this evaluation strategy, classifying one Pets image requires 18 seconds on a RTX 3090 GPU. As our work focuses on understanding diffusion model capabilities and not on developing a fully practical inference algorithm, we do not significantly tune the evaluation strategies. Further details on adaptive evaluation are in \cref{sec:accel_algo}. 

Further reducing inference time could be a valuable avenue for future work. 
Inference is still impractical when there are many classes. Classifying a single ImageNet image, with 1000 classes, takes about 1000 seconds with Stable Diffusion at $512 \times 512$ resolution, even with our adaptive strategy. Table~\ref{tab:prune_time} shows inference times for each dataset, and we discuss promising approaches for speedups in \cref{sec:discussion}.

\section{Experimental Details}
We provide setup details, baselines, and datasets for zero-shot and supervised classification.

\begin{table*}[t]
    \centering
        \begin{adjustbox}{width=0.92\textwidth}
    \begin{tabular}{lcccccccccccccccccccc}
    \toprule
       & Zero-shot? & Food & CIFAR10 & Aircraft & Pets & Flowers & STL10 & ImageNet & ObjectNet \\
    \midrule
    Synthetic SD Data             & \greencheck & 12.6 & 35.3 & 9.4 & 31.3 & 22.1 & 38.0 & 18.9 & 5.2 \\
    \color{dt}{SD Features}        &     \xmark       & \color{dt}{73.0} & \color{dt}{84.0} & \color{dt}{\textbf{35.2}} & \color{dt}{75.9} & \color{dt}{\textbf{70.0}} & \color{dt}{87.2} & \color{dt}{56.6} & \color{dt}{10.2} \\
    Diffusion Classifier (ours)          & \greencheck & \textbf{77.7}   &\bf{88.5}& 26.4 &  \bf{87.3}  & 66.3  &\bf{95.4}& \bf{61.4}& \bf{43.4} \\
    \midrule
    CLIP ResNet-50                              & \greencheck & 81.1 & 75.6 & 19.3 & 85.4 & 65.9 & 94.3 & 58.2 & 40.0 \\
    OpenCLIP ViT-H/14                           & \greencheck & 92.7 & 97.3 & 42.3 & 94.6 & 79.9 & 98.3 & 76.8 & 69.2 \\

    \bottomrule
    \vspace{-1mm}
    \end{tabular}
    \end{adjustbox}
    \caption{\textbf{Zero-shot classification performance}. Our zero-shot \model method (which utilizes Stable Diffusion) significantly outperforms the zero-shot diffusion model baseline that trains a classifier on synthetic SD data. Diffusion Classifier also generally outperforms the baseline trained on Stable Diffusion features, despite ``SD Features'' using the entire training set to train a classifier. Finally, although making a fair comparison is difficult due to different training datasets, our generative approach surprisingly outperforms CLIP ResNet-50 and is competitive with OpenCLIP ViT-H. We report average accuracy or mean-per-class accuracy in accordance with \cite{kornblith2019better}.
    }
    \label{tab:zero_shot_cls}
\end{table*}

\subsection{Zero-shot Classification}

\paragraph{Diffusion Classifier Setup:} Zero-shot \model utilizes Stable Diffusion 2.0~\cite{rombach2022high}, a text-to-image latent diffusion model trained on a filtered subset of LAION-5B~\cite{schuhmann2022laion}. 
Additionally, instead of using the squared $\ell_2$ norm to compute the $\epsilon$-prediction error, we leave the choice between $\ell_1$ and $\ell_2$ as a per-dataset inference hyperparameter. See \cref{sec:loss_fn} for more discussion. We also use the adaptive Diffusion Classifier from Algorithm~\ref{alg:diffusion_classifier_accel}.

\vspace{-0.15in}
\paragraph{Baselines:} We provide results using two strong discriminative zero-shot models: (a) CLIP ResNet-50 \cite{radford2021learning} 
and (b) OpenCLIP ViT-H/14 \cite{cherti2022reproducible}. 
We provide these for reference only, as these models are trained on different datasets with very different architectures from ours and thus cannot be compared apples-to-apples.
We further compare our approach against two alternative ways to extract class labels from diffusion models: (c) \textbf{Synthetic SD Data}: We train a ResNet-50 classifier on synthetic data generated using Stable Diffusion (with class names as prompts), (d) \textbf{SD Features}: This baseline is not a zero-shot classifier, as it requires a \textbf{labeled dataset} of real-world images and class-names. Inspired by Label-DDPM \cite{baranchuk2022labelefficient}, we extract Stable Diffusion features (mid-layer U-Net features at a resolution [$8 \times 8 \times 1024$] at timestep $t=100$), and then fit a ResNet-50 classifier on the extracted features and corresponding ground-truth labels. Details are in \cref{sec:baseline_details}.

\vspace{-0.15in}
\paragraph{Datasets:} We evaluate the zero-shot classification performance across eight datasets: Food-101 \cite{bossard14}, CIFAR-10 \cite{CIFAR-10},  FGVC-Aircraft \cite{maji13fine-grained}, Oxford-IIIT Pets \cite{parkhi12a}, Flowers102 \cite{Nilsback08}, 
STL-10 \cite{pmlr-v15-coates11a}, ImageNet \cite{deng2009imagenet}, and ObjectNet \cite{Barbu2019ObjectNetAL}. 
Due to computational constraints, we evaluate on 2000 test images for ImageNet. We also evaluate zero-shot compositional reasoning ability on the Winoground benchmark~\cite{thrush2022winoground}.

\subsection{Supervised Classification}
\paragraph{\model Setup:} We build \model on top of Diffusion Transformer (DiT) \cite{Peebles2022DiT}, a class-conditional latent diffusion model trained only on ImageNet-1k \cite{deng2009imagenet}. We use DiT-XL/2 at resolution $256^2$ and $512^2$ and evaluate each class 250 times per image.

\vspace{-0.15in}
\paragraph{Baselines:} We compare against the following discriminative models trained with cross-entropy loss on ImageNet-1k: ResNet-18, ResNet-34, ResNet-50, and ResNet-101~\cite{he2016deep}, as well as ViT-L/32, ViT-L/16, and ViT-B/16~\cite{dosovitskiy2020image}.

\vspace{-0.15in}
\paragraph{Datasets:} We evaluate models on their in-distribution accuracy on ImageNet \cite{deng2009imagenet} and out-of-distribution generalization to ImageNetV2 \cite{Recht2019DoIC}, ImageNet-A \cite{hendrycks2021nae}, and ObjectNet \cite{Barbu2019ObjectNetAL}. ObjectNet accuracy is computed on the 113 classes shared with ImageNet. Due to computational constraints, we evaluate Diffusion Classifier accuracy on 10,000 validation images for ImageNet. We compute the baselines' ImageNet accuracies on the same 10,000 image subset.

\section{Experimental Results}
In this section, we conduct detailed experiments aimed at addressing the following questions:
\begin{enumerate}[noitemsep,topsep=0pt]
    \item How does Diffusion Classifier compare against zero-shot state-of-the-art classifiers such as CLIP?
    \item How does our method compare against alternative approaches for classification with diffusion models?
    \item How well does our method do on compositional reasoning tasks?
    \item How well does our method compare to discriminative models trained on the same dataset?
    \item How robust is our model compared to discriminative classifiers over various distribution shifts?
\end{enumerate}

\subsection{Zero-shot Classification Results}
\label{subsec:zero_shot}
Table \ref{tab:zero_shot_cls} shows that \model significantly outperforms the Synthetic SD Data baseline, an alternate zero-shot approach of extracting information from diffusion models. This is likely because the model trained on synthetically generated data learns to rely on features that do not transfer to real data. Surprisingly, our method also generally outperforms the SD Features baseline, which is a classifier trained in a \textit{supervised} manner using the entire \textit{labeled training set} for each dataset. In contrast, our method is zero-shot and requires no additional training or labels. Finally, while it is difficult to make a fair comparison due to training dataset differences, our method outperforms CLIP ResNet-50 and is competitive with OpenCLIP ViT-H. 

This is a major advancement in the performance of generative approaches, and there are clear avenues for improvement. First, we performed no manual prompt tuning and simply used the prompts used by the CLIP authors. Tuning the prompts to the Stable Diffusion training distribution should improve its recognition abilities. 
Second, we suspect that Stable Diffusion classification accuracy could improve with a wider training distribution. 
Stable Diffusion was trained on a subset of LAION-5B \cite{schuhmann2022laion} filtered aggressively to remove low-resolution, potentially NSFW, or unaesthetic images. This decreases the likelihood that it has seen relevant data for many of our datasets. The rightmost column in Table~\ref{tab:sd_filter} shows that only 0-3\% of the test images in CIFAR10, Pets, Flowers, STL10, ImageNet, and ObjectNet would remain after applying all three filters. \textit{Thus, many of these zero-shot test sets are completely out-of-distribution for Stable Diffusion.} Diffusion Classifier performance would likely improve significantly if Stable Diffusion were trained on a less curated training set.

\begin{table}[]
    \centering
    \begin{adjustbox}{width=\linewidth}
    \begin{tabular}{lccccc}
    \toprule
    Dataset & Resolution & Aesthetic & SFW & A + S & R + A + S \\
    \midrule
    Food & 61.5 & 90.5 & 99.9 & 90.5 & 56.3 \\
    CIFAR10 & 0.0 & 3.4 & 90.3 & 3.2 & 0.0 \\
    Aircraft & 98.6 & 95.7 & 100.0 & 95.6 & 94.4 \\
    Pets & 1.1 & 89.1 & 100.0 & 89.1 & 0.9 \\
    Flowers & 0.0 & 82.4 & 100.0 & 82.4 & 0.0 \\
    STL10 & 0.0 & 31.6 & 93.1 & 30.6 & 0.0 \\
    ImageNet & 4.5 & 84.1 & 98.0 & 82.5 & 3.4 \\
    ObjectNet & 98.8 & 20.5 & 98.8 & 20.3 & 20.2 \\
\bottomrule
    \end{tabular}
    \end{adjustbox}
    \caption{\textbf{How in-distribution is each test set for Stable Diffusion?} 
    We show the percentage of each test set that would remain after the Stable Diffusion 2.0 data filtering process. The first three columns show the percentage of images that pass resolution ($\geq 512^2$), aesthetic ($\geq$ 4.5), and safe-for-work ($\leq 0.1$) thresholds, respectively. The last two columns show the proportion of images that pass multiple filters, and the last column (R + A + S) corresponds to the actual filtering criteria used to train SD 2.0.}
    \label{tab:sd_filter}
    \vspace{-0.1in}
\end{table}

\subsection{Improved Compositional Reasoning Abilities}

Large text-to-image diffusion models are capable of generating samples with impressive compositional generalization. 
In this section, we test whether this generative ability translates to improved compositional \textit{reasoning}.

\paragraph{Winoground Benchmark:}
We compare Diffusion Classifier to contrastive models like CLIP~\cite{radford2021learning} on Winoground~\cite{thrush2022winoground}, 
a popular benchmark for evaluating the visio-linguistic compositional reasoning abilities of vision-language models. 
Each example in Winoground consists of 2 (image, caption) pairs. %
Notably, both captions within an example contain the exact same set of words, just in a different order. 
Vision-language multimodal models are scored on Winoground by their ability to match captions $C_i$ to their corresponding images $I_i$. 
Given a model that computes a score for each possible pair $score(C_i, I_j)$, the \textit{text score} of a particular 
example $((C_0, I_0), (C_1, I_1))$ is 1 if and only if it independently prefers caption $C_0$ over caption $C_1$ for
 image $I_0$ and vice-versa for image $I_1$. Precisely, the model's text score on an example is: 
\begin{align}
\label{eq:text_score}
\begin{split}
\mathbb{I}[&score(C_0, I_0) > score(C_1, I_0)  \text{ AND } \\
&score(C_1, I_1) > score(C_0, I_1)]
\end{split}
\end{align}
Achieving a high text score is extremely challenging. Humans (via Mechanical Turk) achieve 89.5\% accuracy on this benchmark, 
but even the best models do barely above chance. 
Models can only do well if they understand compositional structure within each modality. 
CLIP has been found to do poorly on this benchmark since its embeddings tend to be more like a ``bag of concepts'' that fail to bind subjects to attributes or verbs~\cite{yamada2022lemons}. 

\begin{figure}[t]
    \centering
    \includegraphics[width=\linewidth]{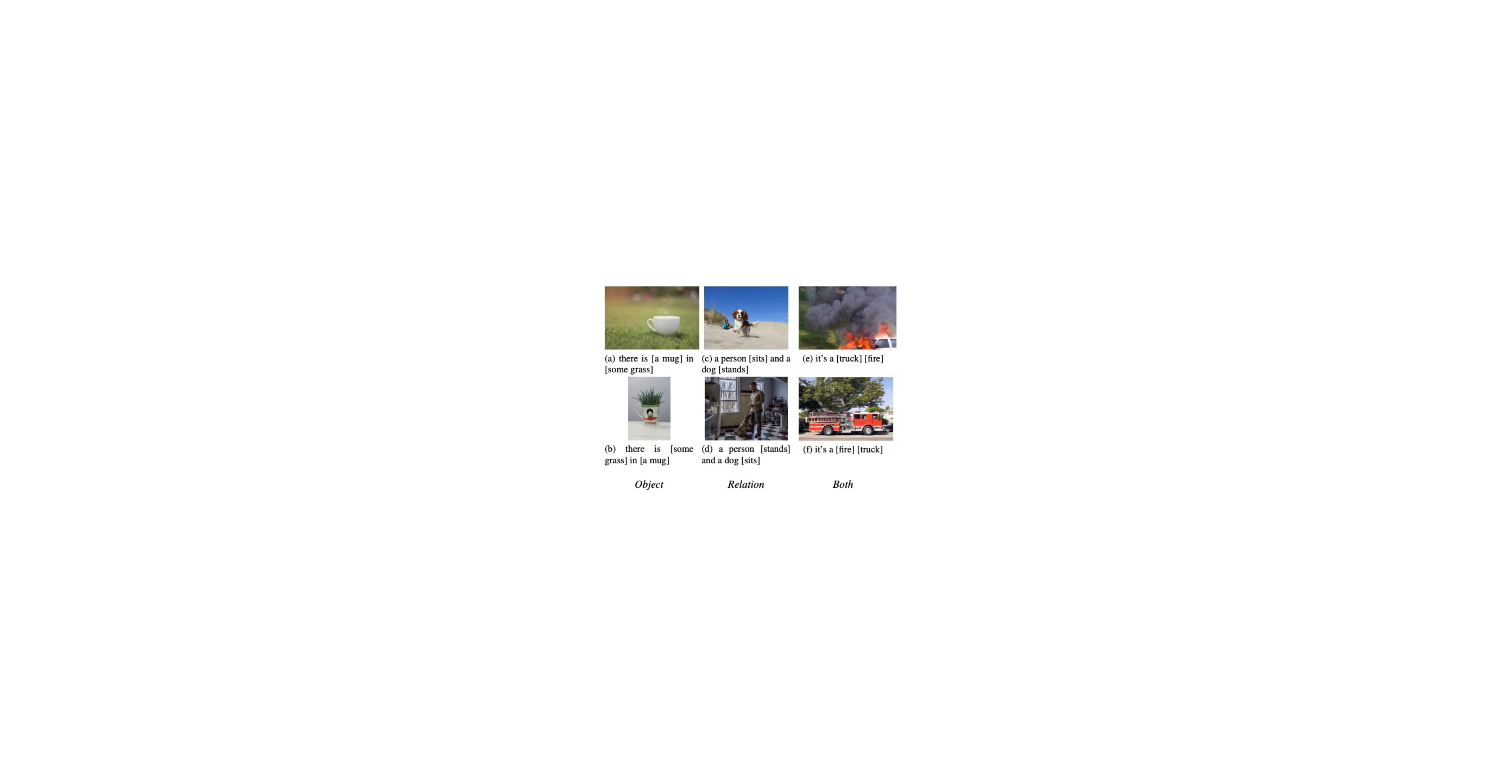}
    \vspace{-0.2in}
    \caption{\textbf{Example visualizations of Winoground swap types.} Each category corresponds to a different type of linguistic swap in the caption. Object swaps noun phrases, Relation swaps verbs, adjectives, or adverbs, and Both can swap entities of both kinds.}
    \label{fig:winoground_explanation}
\end{figure}

Each example is tagged by the type of linguistic swap (object, relation and both) between the two captions: 
\begin{enumerate}[noitemsep,topsep=0pt]
    \item Object: reorder elements like noun phrases that typically refer to real-world objects/subjects.
    \item Relation: reorder elements like verbs,  adjectives,  prepositions,  and/or adverbs that modify objects. 
    \item Both: a combination of the previous two types.
\end{enumerate}
We show examples of each swap type in Figure~\ref{fig:winoground_explanation}.

\begin{table}%
    \centering
    \scalebox{0.85}{
    \begin{tabular}{lcccc}
        \toprule
        Model & Object & Relation & Both & Average \\
        \midrule
        Random Chance & 25.0 & 25.0 & 25.0 & 25.0 \\
        CLIP ViT-L/14 & 27.0 &  25.8 & 57.7 & 28.2 \\
        OpenCLIP ViT-H/14 & 39.0 & 26.6 & 57.7 & 33.0 \\ 
        Diffusion Classifier (ours) & \textbf{46.1} & \textbf{29.2} & \textbf{80.8} & \textbf{38.5} \\
        \bottomrule
    \end{tabular}}
    \vspace{1mm}
    \caption{\textbf{Compositional reasoning results on Winoground}. Diffusion Classifier obtains signficantly better text score (Eq.~\ref{eq:text_score}) than the contrastive baselines for all three swap categories. }
    \label{tab:winoground}
    \vspace{-3mm}
\end{table}

\paragraph{Results} 
Table~\ref{tab:winoground} compares Diffusion Classifier to two strong contrastive baselines: OpenCLIP ViT-H/14 (whose text embeddings Stable Diffusion conditions on) and CLIP ViT-L/14. 
\textit{Diffusion Classifier significantly outperforms both discriminative approaches on Winoground}. Our method is stronger on all three swap types, even the challenging ``Relation'' swaps where the contrastive baselines do no better than random guessing. This indicates that Diffusion Classifier's generative approach exhibits better compositional reasoning abilities. Since Stable Diffusion uses the same text encoder as OpenCLIP ViT-H/14, this improvement comes from better cross-modal binding of concepts to images. 
Overall, we find it surprising that Stable Diffusion, trained with only sample generation in mind, can be repurposed into such a strong classifier and reasoner without any additional training.

\subsection{Supervised Classification Results}
\label{sec:supervised}
We compare \model, leveraging the ImageNet-trained DiT-XL/2 model~\cite{Peebles2022DiT}, to ViTs~\cite{dosovitskiy2020image} and ResNets~\cite{he2016deep} trained on ImageNet. This setting is particularly interesting because it enables a fair comparison between models trained on the same dataset. Table~\ref{tab:id_vs_ood} shows that \model outperforms ResNet-101 and ViT-L/32. Diffusion Classifier achieves ImageNet accuracies of 77.5\% and 79.1\% at resolutions $256^2$ and $512^2$ respectively. 
\textit{To the best of our knowledge, we are the first to show that a generative model trained to learn $p_\theta (\mathbf{x} \mid \mathbf{c})$ can achieve ImageNet classification accuracy comparable to highly competitive discriminative methods.} 

\begin{table}
    \centering
    \begin{adjustbox}{width=\linewidth}
    \begin{tabular}{@{\extracolsep{4pt}}lcccc@{}}
    \toprule
    \multirow{2}{*}{\textbf{Method}}
    &\multicolumn{1}{c}{\textbf{ID}} 
    &\multicolumn{3}{c}{\textbf{OOD}} \\
    \cmidrule{2-2} \cmidrule{3-5}
                 & IN & IN-V2 & IN-A & ObjectNet \\
    \midrule
    ResNet-18  & \first 70.3 & \first 57.3 & \first 1.1 & \first 27.2 \\
    ResNet-34  & \first 73.8 & \first 61.0 & \first 1.9 & \first 31.6 \\ 
    ResNet-50  & \first 76.7        & \first 63.2 & \first 0.0 & 36.4 \\
    ResNet-101 & \first 77.7        & \first 65.5        & \first 4.7 & 39.1 \\
    ViT-L/32   & \first 77.9        & \first    64.4        & \first 11.9 & \first 32.1 \\
    ViT-L/16   & 80.4        & 67.5        & \first 16.7 & 36.8 \\
    ViT-B/16   & \textbf{81.2}        & \textbf{69.6}        & \first 20.8 & \textbf{39.9} \\
    \midrule
    Diffusion Classifier $256^2$ & 77.5 & 64.6 & 20.0 & 32.1 \\
    Diffusion Classifier $512^2$ & 79.1 & 66.7 & \textbf{30.2} & 33.9\\
    \bottomrule
    \end{tabular}
    \end{adjustbox}
    \caption{\textbf{Standard classification on ImageNet.} We compare \model (using DiT-XL/2 at $256^2$ and $512^2$ resolutions) to discriminative models trained on ImageNet. We highlight cells where \model does better. All models (generative and discriminative) have only been trained on ImageNet. }
    \label{tab:id_vs_ood}
    \vspace{-3mm}
\end{table}

\subsubsection{Better Out-of-distribution Generalization}
\begin{figure}[t]
    \centering
    \includegraphics[width=\linewidth]{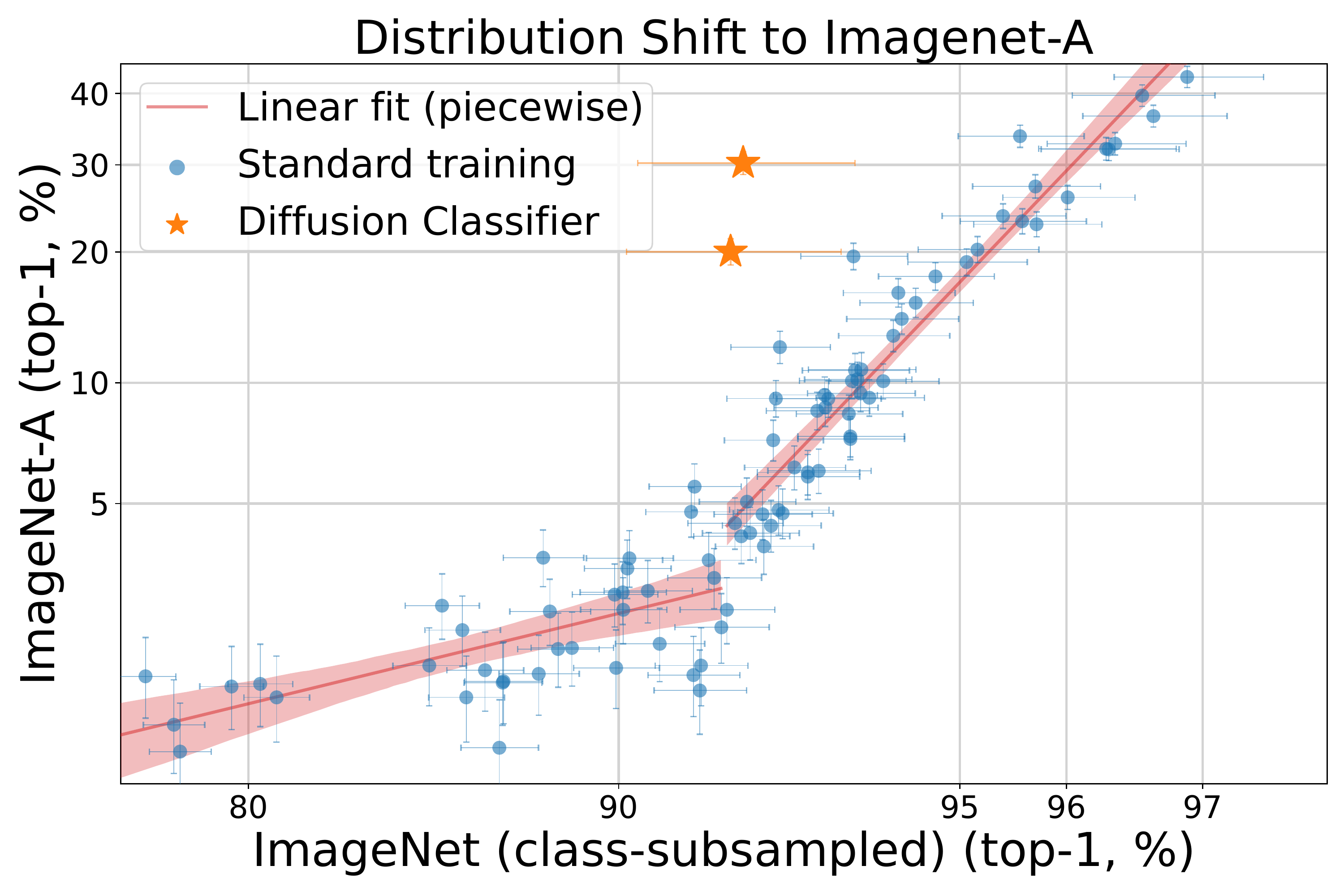}
    \vspace{-4mm}
    \caption{\textbf{Diffusion Classifier exhibits effective robustness without using extra labeled data}.
    Compared to discriminative models trained on the same amount of labeled data (``standard training''), Diffusion Classifier achieves much higher ImageNet-A accuracy than predicted by its ImageNet accuracy. Diffusion Classifier points correspond to DiT-XL/2 at resolution $256^2$ and $512^2$. Points are shown with 99.5\% Clopper-Pearson confidence intervals. The red lines show the linear relationship between ID and OOD accuracy for discriminative models, with a ``break'' at the accuracy of the model used to create ImageNet-A. The axes were adjusted using logit scaling, since accuracies fall within $[0, 100]$.}
    \label{fig:id_vs_ood}
    \vspace{-3mm}
\end{figure}

We find that Diffusion Classifier surprisingly has stronger out-of-distribution (OOD) performance on ImageNet-A than all of the baselines. In fact, our method shows qualitatively different and better OOD generalization behavior than discriminative approaches. Previous work~\cite{taori2020measuring} evaluated hundreds of discriminative models and found a tight linear relationship between their in-distribution (ID) and OOD accuracy --- for a given ID accuracy, no models do better OOD than predicted by the linear relationship. 
For models trained on only ImageNet-1k (no extra data), none of a wide variety of approaches, from adversarial training to targeted augmentations to different architectures, achieve better OOD accuracy than predicted. We show the relationship between ID ImageNet accuracy (subsampled to the classes that overlap with ImageNet-A) and OOD accuracy on ImageNet-A for these discriminative models as the blue points (``standard training'') in Figure~\ref{fig:id_vs_ood}. The OOD accuracy is described well by a piecewise linear fit, with a kink at the ImageNet accuracy of the ResNet-50 model used to identify the hard images that comprise ImageNet-A. No discriminative models show meaningful ``effective robustness,'' which is the gap between the actual OOD accuracy of a model and the OOD accuracy predicted by the linear fit~\cite{taori2020measuring}.

However, in contrast to these hundreds of discriminative models, Diffusion Classifier achieves much higher OOD accuracy on ImageNet-A than predicted. Figure~\ref{fig:id_vs_ood} shows that Diffusion Classifier lies far above the linear fit and achieves an effective robustness of 15-25\%. \textit{To the best of our knowledge, this is the first approach to achieve significant effective robustness without using any extra data during training.}

There are a few caveats to our finding. Diffusion Classifier does not show improved effective robustness on the ImageNetV2 or ObjectNet distribution shifts, though perhaps the nature of those shifts is different from that of ImageNet-A. Diffusion Classifier may do better on ImageNet-A since its predictions could be less correlated with the (discriminative) ResNet-50 used to find hard examples for ImageNet-A. Nevertheless, the dramatic improvement in effective robustness on ImageNet-A is exciting and suggests that generative classifiers are promising approaches to achieve better robustness to distribution shift.

\subsubsection{Stable Training and No Overfitting}
Diffusion Classifier's ImageNet accuracy is especially impressive since DiT was trained with \textit{only random horizontal flips}, unlike typical classifiers that use RandomResizedCrop, Mixup~\cite{zhang2017mixup}, RandAugment~\cite{cubuk2020randaugment}, and other tricks to avoid overfitting. Training DiT with more advanced augmentations should further improve its accuracy. Furthermore, DiT training is stable with fixed learning rate and no regularization other than weight decay \cite{Peebles2022DiT}. This stands in stark contrast with ViT training, which is unstable and frequently suffers from NaNs, especially for large models~\cite{MaskedAutoencoders2021}. These results indicate that the generative objective $\log p_\theta (\mathbf{x} \mid \mathbf{c})$ could be a promising way to scale up training to even larger models without overfitting or instability. 

\subsubsection{Choice of classification objective}
While Stable Diffusion parameterizes $p_\theta(\mathbf{x}_{t-1} \mid \mathbf{x}_t, \mathbf{c})$ as a Gaussian with fixed variance, DiT learns the variance $\Sigma_\theta(\mathbf{x_t}, \mathbf{c})$. 
A single network outputs $\epsilon_\theta$ and $\Sigma_\theta$, but they are trained via two separate losses. $\epsilon_\theta$ is trained via a uniform weighting of $\ell_2$ errors $\mathbb{E}_{\epsilon, t}[\|\epsilon - \epsilon_\theta(\mathbf{x}_t, \mathbf{c})\|^2]$, as this is found to improve sample quality. 
In contrast, $\Sigma_\theta$ is trained with the exact variational lower bound. This keeps the timestep-dependent weighting term $w_t$ in Eq.~\ref{eq:exact_elbo} and weights the $\epsilon$-prediction errors by the inverse of the variances $\Sigma_\theta$ (see \cite{Peebles2022DiT} for more details). Since both losses are used at training time, we run an experiment to see which objective yields the best accuracy as an inference-time objective. Instead of choosing the class with the lowest error based on uniform $\ell_2$ weighting, as is done in Algorithm~\ref{alg:diffusion_classifier_naive}, we additionally try using the variational bound or the sum of the uniform weighting and the variational bound. Table~\ref{tab:DiT_objective} shows that the uniform $\ell_2$ weighting does best across all datasets. This justifies the approximation we made to the ELBO in Eq.~\ref{eq:elbo}. The sum of the uniform $\ell_2$ and the variational bound does almost as well, likely because the magnitude of the variational bound is much smaller than that of the uniformly weighted $\ell_2$, so their sum is dominated by the $\ell_2$ term.

\begin{table}[]
    \centering
    \begin{adjustbox}{width=\linewidth}
    \begin{tabular}{clcccc}
    \toprule
    Resolution & Objective & IN & IN-v2 & IN-A & ObjectNet \\
    \midrule
            & $\ell_2$       & \textbf{77.5} & \textbf{64.6} & \textbf{20.0} & \textbf{33.9} \\ 
    $256^2$ & VLB            & 71.6 & 57.7 & 17.9 & 24.7 \\
            & $\ell_2$ + VLB & \textbf{77.5} & \textbf{64.6} & \textbf{20.0} & 33.8 \\
            
    \midrule 
            & $\ell_2$       & \textbf{79.1} & \textbf{66.7} & \textbf{30.2} & \textbf{33.9} \\ 
    $512^2$ & VLB &            74.0 & 59.1 & 24.9 & 24.7 \\
            & $\ell_2$ + VLB & 79.0 & 66.6 & \textbf{30.2} & 33.8 \\
    \bottomrule
    \end{tabular}
    \end{adjustbox}
    \caption{\textbf{Effect of classification objective}. DiT trains $\epsilon_\theta$ with the uniformly weighted $\ell_2$ loss to evaluate $\sum_t w_t \|\epsilon - \epsilon_\theta(\mathbf{x}_t, \mathbf{c})\|^2$ from Eq.~\ref{eq:exact_elbo}. DiT also trains the learned variance $\Sigma_\theta$ of $p_\theta (\mathbf{x}_{t-1} | \mathbf{x}_t)$ with the exact variational lower bound, which weights timesteps unevenly. Since both of these weightings are involved in DiT training, we try each objective, as well as their sum, to see which one achieves the best accuracy. We find that uniformly weighting $\ell_2$ errors across timesteps performs best. 
    }
    \label{tab:DiT_objective}
\end{table}

\section{Conclusion and Discussion}
\label{sec:discussion}
We investigated the zero-shot and standard classification abilities of diffusion models by leveraging them as conditional density estimators. By performing a simple, unbiased Monte Carlo estimate of the learned conditional ELBO for each class, we extract \textbf{\model}---a \textit{powerful approach to turn any conditional diffusion model into a classifier without any additional training.}
We find that this classifier narrows the gap with state-of-the-art discriminative approaches on zero-shot and standard classification and significantly outperforms them on multimodal compositional reasoning. Diffusion Classifier also exhibits far better ``effective robustness'' to distribution shift. 

\paragraph{Accelerating Inference} While inference time is currently a practical bottleneck, there are several clear ways to accelerate Diffusion Classifier. Decreasing resolution from the default $512 \times 512$ (for SD) would yield a dramatic speedup. Inference at $256\times 256$ is at least $4 \times$ faster, and inference at $128 \times 128$ would be over $16\times$ faster. Another option is to use a weak discriminative model to quickly eliminate classes that are clearly incorrect. Appendix~\ref{sec:prune} shows that this would simultaneously improve accuracy and reduce inference time. Gradient-based search could backpropagate through the diffusion model to solve $\argmax_c \log p(x \mid c)$, which could eliminate the runtime dependency on the number of classes. New architectures could be designed to only use the class conditioning $c$ toward the end of the network, enabling reuse of intermediate activations across classes. Finally, note that the error prediction process is easily parallelizable. With sufficient scaling or better GPUs in the future, all Diffusion Classifier steps can be done in parallel with the \textit{latency of a single forward pass}.

\paragraph{Role of Diffusion Model Design Decisions}
Since we don't change the base diffusion model of \model, the choices made during diffusion training affect the classifier. For instance, Stable Diffusion \cite{rombach2022high} conditions the image generation on the text embeddings from OpenCLIP~\cite{ilharco2021openclip}. 
However, the language model in OpenCLIP is much weaker than open-ended large-language models like T5-XXL \cite{T5Model} because it is only trained on text data available from image-caption pairs, a minuscule subset of total text data on the Internet. Hence, we believe that diffusion models trained on top of T5-XXL embeddings, such as Imagen~\cite{saharia2022photorealistic}, 
should display better zero-shot classification results, but these are not open-source to empirically validate. Other design choices, such as whether to perform diffusion in latent space (e.g. Stable Diffusion) or in pixel space (e.g. DALLE 2), can also affect the adversarial robustness of the classifier and present interesting avenues for future work.

\vspace{0.1in}
In conclusion, while generative models have previously fallen short of discriminative ones for classification, today's pace of advances in generative modeling means that they are rapidly catching up. Our strong classification, multimodal compositional reasoning, and robustness results represent an encouraging step in this direction.

\vspace{2mm}
\noindent \textbf{Acknowledgements} We thank Patrick Chao for helpful discussions and Christina Baek and Rishi Veerapaneni for paper feedback. Stability.AI contributed compute to run some experiments. AL is supported by the NSF GRFP DGE1745016 and DGE2140739. This work is supported by NSF IIS-2024594 and ONR MURI N00014-22-1-2773.

{\small
\bibliographystyle{ieee_fullname}
\bibliography{main}
}
\newpage
\appendix
\onecolumn
\section*{Appendix}

\section{Efficient Diffusion Classifier Algorithm}
\label{sec:accel_algo}
Though
\model works straightforwardly with the procedure described in \cref{alg:diffusion_classifier_naive},
we are interested in speeding up inference as described in
\cref{sec:efficient_classification}.
\cref{alg:diffusion_classifier_accel} shows the efficient Diffusion Classifier procedure that adaptively chooses which classes to continue evaluating. Table~\ref{tab:eval_strategy} shows the evaluation strategy used for each zero-shot dataset. We hand-picked the strategies based on the number of classes in each dataset. Further gains in accuracy may be possible with more evaluations. 

\begin{algorithm}
   \caption{$\texttt{Diffusion Classifier (Adaptive)}$}
   \label{alg:diffusion_classifier_accel}
\begin{algorithmic}[1]
    \STATE {\bfseries Input:} test image $\mathbf{x}$, conditioning inputs $\mathcal{C} = \{\mathbf{c}_i\}_{i=1}^n$ (e.g., text embeddings or class indices), number of stages $N_{\text{stages}}$, list $\texttt{KeepList}$ of number of $\mathbf{c}_i$ to keep after each stage, list $\texttt{TrialList}$ of number of trials done by each stage 
    \STATE Initialize $\texttt{Errors}[\mathbf{c}_i] = \text{list}()$ for each $\mathbf{c}_i$ 
    \STATE Initialize $\texttt{PrevTrials}$ = 0 \;\; \COMMENT{How many times we've tried each remaining element of $\mathcal C$ so far}
    \FOR{stage i $=1, \dots, N_{\text{stages}}$}
        \FOR{trial $j = 1, \dots, \texttt{TrialList}[i] - \texttt{PrevTrials} $}
            \STATE Sample $t \sim [1, 1000]$
            \STATE Sample $\epsilon \sim \mathcal{N}(0, I)$
            \STATE $\mathbf{x}_t = \sqrt{\bar \alpha_{t}}\mathbf{x} + \sqrt{1-\bar\alpha_{t}} \epsilon$
            \FOR{conditioning $\mathbf{c}_k \in \mathcal{C}$}
                \STATE $\texttt{Errors}[\mathbf{c}_k].\texttt{append}(\|\epsilon - \epsilon_\theta(\mathbf{x}_t, \mathbf{c}_k)\|^2)$
            \ENDFOR
        \ENDFOR
        \STATE $\displaystyle \mathcal{C} \leftarrow \argmin_{\substack{\mathcal S \subset \mathcal{C}; \\ |\mathcal S| = \texttt{KeepList}[i]}} \sum_{c_k \in \mathcal S} \texttt{mean}(\texttt{Errors}[\mathbf{c}_k])$ \;\; \COMMENT{Keep top $\texttt{KeepList}[i]$ conditionings $\mathbf{c}_k$ with the lowest errors} 
        \STATE $\texttt{PrevTrials}$ = \texttt{TrialList}[i]
    \ENDFOR
    \RETURN $\displaystyle \argmin_{\mathbf{c}_i \in \mathcal C} \texttt{\,mean}(\texttt{Errors}[\mathbf{c}_i])$
\end{algorithmic}
\end{algorithm}

\begin{table}[h]
    \centering
    \begin{tabular}{lllcc}
    \toprule
        Dataset & Prompts kept per stage & Evaluations per stage &  Avg. evaluations per class & Total evaluations \\
    \midrule
        Food101 & 20 10 5 1 & 20 50 100 500 & 50.7 & 5120 \\
        CIFAR10 & 5 1 & 50 500 & 275 & 2750 \\
        Aircraft & 20 10 5 1 & 20 50 100 500 & 51 & 5100 \\
        Pets & 5 1 & 25 250 & 51 & 1890\\
        Flowers102 & 20 10 5 1 & 20 50 100 500 & 50.4 & 5140 \\
        STL10 & 5 1 & 100 500 & 300 & 3000 \\
        ImageNet & 500 50 10 1 & 50 100 500 1000 & 100 & 100000\\
        ObjectNet & 25 10 5 1 & 50 100 500 1000 & 118.6 & 13400 \\
        \bottomrule
    \end{tabular}
    \vspace{2mm}
    \caption{Adaptive evaluation strategy for each zero-shot dataset.}
    \label{tab:eval_strategy}
\end{table}

\section{Inference Costs and Hybrid Classification Approach}
\label{sec:prune}
\begin{table}[h]
    \centering
        \begin{adjustbox}{width=1\linewidth}
    \begin{tabular}{lccccccc}
    \toprule
       & Food101 & CIFAR10 & Aircraft & Oxford Pets & Flowers102 & STL10 & ImageNet \\
    \midrule
    Diffusion Classifier                 &  77.7 & 88.5          & 26.4 & 87.3          & 66.3           & 95.4 & 61.4 \\
    \;\; \color{dt}{Time/img (s)} & \color{dt}{51} & \color{dt}{30} & \color{dt}{51} & \color{dt}{18} & \color{dt}{51} & \color{dt}{30} & \color{dt}{1000} \\
    Diffusion Classifier w/ discriminative pruning & 78.7 & 88.5 & 26.8 & 86.4 & 67.0 & 95.4 & 62.6 \\
    \;\; \color{dt}{Time/img (s)} & \color{dt}{35} & \color{dt}{30} & \color{dt}{35} & \color{dt}{18} & \color{dt}{35} & \color{dt}{30} & \color{dt}{150} \\
    \;\; 
    \color{dt}{Est. Time/img (s) at $128^2$ res}
    & \color{dt}{2} & \color{dt}{2} & \color{dt}{2} & \color{dt}{1} & \color{dt}{2} & \color{dt}{2} & \color{dt}{9} \\
    
    \midrule
    CLIP ResNet-50                              & 81.1 & 75.6 & 19.3 & 85.4 & 65.9 & 94.3 & 58.2 \\
    
    \bottomrule
    \vspace{-2mm}
    \end{tabular}
    \end{adjustbox}
    \caption{\textbf{Zero-shot accuracy and inference time with Stable Diffusion $512 \times 512$.} ``Pruning'' away unlikely classes with a weak discriminative classifier (e.g., CLIP ResNet-50) increases accuracy and reduces inference time. Additionally, reducing resolution to $128 \times 128$ would reduce inference time by roughly $16\times$. However, its impact on accuracy is difficult to estimate without retraining the Stable Diffusion model to expect lower resolutions. All times are estimated using a RTX 3090 GPU.  }
    \label{tab:prune_time}
\end{table}

Table~\ref{tab:prune_time} shows the inference time of Diffusion Classifier when using the efficient Diffusion Classifier algorithm (Algorithm~\ref{alg:diffusion_classifier_accel}). Classifying a single image takes anywhere between 18 seconds (Pets) to 1000 seconds (ImageNet). The issue with ImageNet is that Diffusion Classifier inference time still approximately scales linearly with the number of classes, even when using the adaptive strategy. One way to address this problem is to use a weak discriminative model to quickly ``prune'' away classes that are almost certainly incorrect. Table~\ref{tab:prune_time} shows that using Diffusion Classifier to choose among the top 20 class predictions made by CLIP ResNet-50 for an image greatly reduces inference time, while even improving performance. This pruning procedure only requires the top-20 accuracy of the fast discriminative model to be high (close to 100\%), so it works even when the top-1 accuracy of the ResNet-50 is low, like on Aircraft. We chose top-20 intuitively, without any hyperparameter search, and tuning the $k$ for top-$k$ pruning will trade off between inference time and accuracy. Note that no other results in this paper use the discriminative pruning procedure, to avoid conflating the capabilities of Diffusion Classifier with those of the weak discriminative model used to prune. 

\section{Inference Objective Function}
\label{sec:loss_fn}

\begin{table*}[h]
    \centering
        \begin{adjustbox}{width=0.95\textwidth}
    \begin{tabular}{lccccccccccccccccccc}
    \toprule
                     & Food101       & CIFAR10       & Aircraft      & Oxford Pets   & Flowers102     & STL10         & ImageNet      & ObjectNet \\
    \midrule
    Squared $\ell_2$ & \textbf{77.7} & 84.4          & 26.4          & 86.3          & 62.2           & \textbf{95.4} & \textbf{61.4} & 43.4 \\
    $\ell_1$         & 73.8          & \textbf{88.4} & 22.1          & \textbf{87.3} & \textbf{66.3}  & \textbf{95.4} & 59.6          &  36.8 \\
    Huber            & \textbf{77.7} & 84.6          & \textbf{26.7} & 86.6          & 62.6           & \textbf{95.4} & 60.9          & \textbf{43.5} \\
    \bottomrule
    \vspace{-2mm}
    \end{tabular}
    \end{adjustbox}
    \caption{\textbf{Diffusion Classifier zero-shot performance with different loss functions $\mathcal{L}(\epsilon - \epsilon_\theta(\mathbf{x}_t, \mathbf{c}))$}. 
    }
    \label{tab:loss_fn}
\end{table*}

\begin{table*}[h]
    \centering
    \begin{tabular}{clcccc}
     \toprule 
       Resolution & Objective & ImageNet & ImageNetV2 & ImageNet-A & ObjectNet \\
      \midrule
      $256^2$ & Squared $\ell_2$ & \textbf{77.5} & \textbf{64.6} & \textbf{20.0} & \textbf{32.1} \\
      $256^2$ & $\ell_1$         & 74.9 & 60.5 & 9.7 & 24.7 \\
      \midrule 
      $512^2$ & Squared $\ell_2$ & \textbf{79.1} & \textbf{66.7} & \textbf{30.2} & \textbf{33.9} \\
      $512^2$ & $\ell_1$ & 75.6 & 62.1 & 13.2 & 26.2 \\
      \bottomrule
    \end{tabular}
    \vspace{2mm}
    \caption{\textbf{Diffusion Classifier supervised performance with different loss functions $\mathcal{L}(\epsilon - \epsilon_\theta(\mathbf{x}_t, \mathbf{c}))$}. 
    }
    \label{tab:dit_loss_fn}
\end{table*}

While the theory in Section~\ref{sec:method} justifies using $\|\epsilon - \epsilon_\theta(\mathbf{x}_t, \mathbf{c})\|_2^2$ within the Diffusion Classifier inference objective, we surprisingly find that other loss functions can work better in some cases. 
Table~\ref{tab:loss_fn} shows that $\|\epsilon - \epsilon_\theta(\mathbf{x}_t, \mathbf{c})\|_1$ (the $\ell_1$ loss) instead of the squared $\ell_2$ loss does better on roughly half of the datasets that we use to evaluate the Stable Diffusion-based zero-shot classifier. This is puzzling, since the $\ell_1$ loss is neither theoretically justified nor appears in the Stable Diffusion training objective. We hope followup work can explain the empirical success of the $\ell_1$ loss. 
Combining these two losses does not get the ``best of both worlds.'' The Huber loss, which is the squared $\ell_2$ loss for values less than 1 and is the $\ell_1$ loss for values greater than 1, roughly achieves the same performance as the theoretically-justified squared $\ell_2$ loss. We choose between squared $\ell_2$ and $\ell_1$ as a hyperparameter for Section~\ref{subsec:zero_shot}. Table~\ref{tab:dit_loss_fn} shows that $\ell_1$ does not help with supervised classification (Section~\ref{sec:supervised}) using DiT-XL/2.

\section{Interpretability via Image Generation}
In contrast to discriminative classifiers, where it is difficult to understand what features the model has learned or why a model has made a certain decision, generative classifiers are easier to visualize. In this section, we examine how samples from the generative model can help us understand class-dependent features that the model has learned as well as failures in the model's understanding. 

\begin{figure*}[t]
    \centering
    \includegraphics[width=0.8\linewidth]{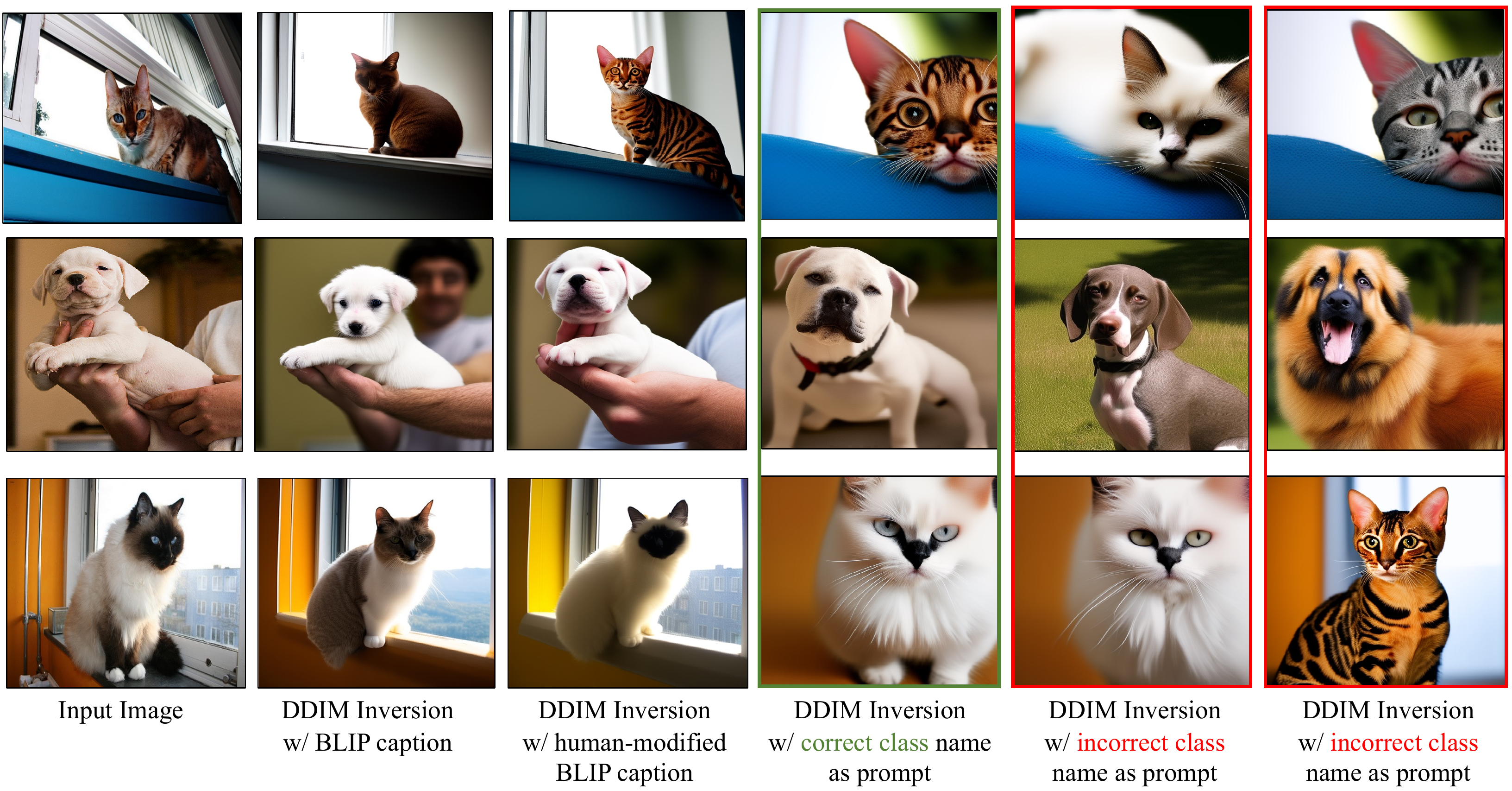}
    \caption{\textbf{Analyzing \model for Zero-Shot Classification:} We analyze the role of different text/captions (BLIP, Human-modified BLIP, correct class-name, incorrect class-name) for zero-shot classification using text-based diffusion models. To do so, we invert the input image using the corresponding caption and then reconstruct it using deterministic DDIM sampling. The image inverted and reconstructed using a human-modified BLIP caption aligns the most with the input image since this caption is the most descriptive. The images reconstructed using \colorbox{green!15}{correct class names as prompts (column 4)} align much better with the input image in terms of class-descriptive features of the underlying object than the images reconstructed using \colorbox{red!15}{incorrect class names as prompts (columns 5 and 6)}. Row 3 (columns 4 and 5) demonstrates an example where the base Stable Diffusion does not understand the difference between the two cat breeds, Birman and Ragdoll, and hence cannot invert/sample them differently. As a result, our classifier also fails.}
    \label{fig:zero_shot_analysis}
\end{figure*}

\paragraph{Experiment Setup} Given an input image, we first perform DDIM inversion \cite{DDIM, Kim_2022_CVPR} (with 50 timesteps) using Stable Diffusion 2.0 and different captions as prompts: BLIP \cite{li2022blip} generated caption, human-refined BLIP generated caption, ``a photo of \textit{\{correct-class-name\}}, a type of pet'' and ``a photo of \textit{\{incorrect-class-name\}}, a type of pet.''.  Next, we leverage the inverted DDIM latent and the corresponding prompt to attempt to reconstruct the original image (using a deterministic diffusion scheduler \cite{DDIM}). 
The underlying intuition behind this experiment is that the inverted image should look more similar to the original image when a correct and appropriate/descriptive prompt is used for DDIM inversion and sampling.

\paragraph{Experimental Evaluation} Figure~\ref{fig:zero_shot_analysis} shows the results of this experiment for the Oxford-IIIT Pets dataset. The image inverted using a human-modified BLIP caption (column 3) is the most similar to the original image (column 1). This aligns with our intuition as this caption is most descriptive of the input image. The human-modified caption only adds the correct class name (Bengal Cat, American Bull Dog, Birman Cat) ahead of the BLIP predicted ``cat or dog'' token for the foreground object and slightly enhances the description for the background. 
Comparing the BLIP-caption results (column 2) with the human-modified BLIP-caption results (column 3), we can see that by just using the class-name as the extra token, the diffusion model can inherit class-descriptive features. The Bengal cat has stripes, the American Bulldog has a wider chin, and the Birman cat has a black patch on its face in the reconstructed image. 

Compared to the images generated using the human-generated caption as a prompt, the images reconstructed using only class names as prompts (columns 4,5,6) align less with the input image (column 1). This is expected, as class names by themselves are not dense descriptions of the input images. Comparing the results of column 4 (correct class names as prompt) with those of column 5,6 (incorrect class names as prompt), we can see that the foreground object has similar class-descriptive features (brown and black stripes in row 1 and black face patches in row 3) to the input image for the correct-prompt reconstructions. This highlights the fact that although using class names as approximate prompts will not lead to perfect denoising (Eq.~\ref{eq:monte_carlo}), \textit{for the global prediction task of classification, the correct class names should provide enough descriptive features for denoising, relative to the incorrect class names.} 

Row 3 of Figure~\ref{fig:zero_shot_analysis} further highlights an example of a failure mode where Stable Diffusion generates very similar inverted images for correct Birman and incorrect Ragdoll text prompts. As a result, our model also incorrectly classifies the Birman cat as a Ragdoll. 
To fix this failure mode, we tried finetuning the Stable Diffusion model on a dataset of Ragdoll/Birman cats (175 images in total). Using this finetuned model, Diffusion Classifier accuracy on these two classes increases to 85\%, from an initial zero-shot accuracy of 45\%. In addition to minimizing the standard $\epsilon$-prediction error $\|\epsilon - \epsilon_\theta(\mathbf{x}_t, \mathbf{c}_i)\|^2$, we found that adding a loss term to \textit{increase} the error $\|\epsilon - \epsilon_\theta(\mathbf{x}_t, \mathbf{c}_j)\|^2$ for the wrong class $\mathbf{c}_j$ helped the model distinguish these commonly-confused classes.

\section{How Does Stable Diffusion Version Affect Zero-Shot Accuracy?}

\begin{table}[h]
    \centering
        \begin{adjustbox}{width=0.85\linewidth}
    \begin{tabular}{ccccccccc}
    \toprule
      SD Version & Food101 & CIFAR10 & Aircraft & Oxford Pets & Flowers102 & STL10   & ImageNet & ObjectNet \\
    \midrule
    1.1          & 60.3    & 83.4    & 20.1     &    78.8     & 43.1       & 92.6    & 51.7     & 38.1      \\
    1.2          & 75.7    & 85.9    & 26.3     &    85.4     & 54.4       & 94.4    & 57.3     & 39.4      \\
    1.3          & 77.5    &\bf{87.5}& 27.8     &    87.2     & 54.5       &\bf{94.9}& \bf{59.7}& 40.9      \\
    1.4          & 77.8    & 86.0    & 28.6     &    87.4     & 54.2       & 94.8    & 59.2     & 41.2      \\
    1.5          &\bf{78.4}& 85.5    & \bf{29.1}&  \bf{87.5}  & \bf{55.0}  & 94.5    & 59.6     & \bf{41.6} \\
    \midrule 
    2.0          &  77.7   &\bf{88.5}& \bf{26.4}&  \bf{87.3}  & \bf{66.3}  &\bf{95.4}& \bf{61.4}& \bf{43.4} \\
    2.1          &\bf{77.9}& 87.1    & 24.3     &    86.2     & 59.4       &95.3& 58.4     & 38.3 \\
    \bottomrule
    \vspace{-2mm}
    \end{tabular}
    \end{adjustbox}
    \caption{\textbf{Effect of Stable Diffusion version on Diffusion Classifier zero-shot accuracy}. We bold the best version within SD 1.x and 2.x. For SD 1, accuracy tends to increase with more training. The main exception is on low-resolution datasets like CIFAR10 and STL10. SD 2 performance consistently decreases from SD 2.0 to SD 2.1 on almost every dataset.}
    \label{tab:sd_version}
\end{table}

We investigate how much the Stable Diffusion checkpoint version affects Diffusion Classifier's zero-shot classification accuracy. Table~\ref{tab:sd_version} shows zero-shot accuracy for each Stable Diffusion release version so far. We use the same adaptive evaluation strategy (Algorithm~\ref{alg:diffusion_classifier_accel}) for each version. Accuracy improves with each new release for SD 1.x, as more training likely reduces underfitting on the training data. However, accuracy actually decreases when going from SD 2.0 to 2.1. The cause of this is not clear, especially without access to intermediate training checkpoints. One hypothesis is that further training on $512^2$ resolution images causes the model to forget knowledge from its initial $256^2$ resolution training set, which is closer to the distribution of these zero-shot benchmarks. SD 2.1 was finetuned using a more permissive NSFW threshold ($\geq 0.98$ instead of $\geq 0.1$), so another hypothesis is that this introduced a lot of human images that hurt performance on our object-centric benchmarks.

\section{Additional Implementation Details}

\subsection{Zero-shot classification using Diffusion Classifier}
\label{sec:zero_shot_details}
\paragraph{Training Data}
For our zero-shot Diffusion Classifier, we utilize Stable Diffusion 2.0~\cite{rombach2022high}. 
This model was trained on a subset of the LAION-5B dataset, filtered so that the training data is aesthetic and appropriately safe-for-work. 
LAION classifiers were used to remove samples that are too small (less than $512 \times 512$), potentially not-safe-for-work (punsafe $\geq 0.1$), or unaesthetic (aesthetic score $\leq 4.5$). 
These thresholds are conservative, since false negatives (NSFW or undesirable images left in the training set) 
are worse than removing extra images from a large starting dataset. 
As discussed in Section~\ref{subsec:zero_shot}, these filtering criteria bias the distribution of Stable Diffusion training data 
and likely negatively affect Diffusion Classifier's performance on datasets whose images do not satisfy these criteria. 
SD 2.0 was trained for 550k steps at resolution $256 \times 256$ on this subset, 
followed by an additional 850k steps at resolution $512 \times 512$ on images that are at least that large. 
This checkpoint can be downloaded online through the diffusers repository at 
\verb|stabilityai/stable-diffusion-2-0-base|.

\paragraph{Inference Details}
We use FP16 and Flash Attention~\cite{dao2022flashattention} to improve inference speed. 
This enables efficient inference with a batch size of 32, which works across a variety of GPUs, from RTX 2080Ti to A6000. 
We found that adding these two tricks did not affect test accuracy compared to using FP32 without Flash Attention. 
Given a test image, we resize the shortest edge to 512 pixels using bicubic interpolation, 
take a $512 \times 512$ center crop, and normalize the pixel values to $[-1, 1]$. 
We then use the Stable Diffusion autoencoder to encode the $512 \times 512 \times 3$ RGB image into a $64 \times 64 \times 4$ latent.
We finally classify the test image by applying the method described in Sections~\ref{sec:method} and \ref{sec:practical} to estimate $\epsilon$-prediction error in this latent space.

\subsection{Compositional reasoning using Diffusion Classifier}
For our experiments on the Winoground benchmark~\cite{thrush2022winoground}, most details are the same as the zero-shot details described in Appendix~\ref{sec:zero_shot_details}. We use Stable Diffusion 2.0, and we evaluate each image-caption pair with 1000 evenly spaced timesteps. We omit the adaptive inference strategy since there are only 4 image-caption pairs to evaluate for each Winoground example.

\subsection{ImageNet classification using Diffusion Classifier}

For this task, we use the recently proposed Diffusion Transformer (DiT)~\cite{Peebles2022DiT} as the backbone of our Diffusion Classifier. DiT was trained on ImageNet-1k, which contains about 1.28 million images from 1,000 classes. 
While it was originally trained to produce high-quality samples with strong FID scores, we repurpose the model and compare it against discriminative models with the same ImageNet-1k training data. We use the DiT-XL/2 model size at resolution $256^2$ and $512^2$. Notably, DiT achieves strong performance while using much weaker data augmentations than what discriminative models are usually trained with. During training time for the $256^2$ checkpoint, the smaller edge of the input image is resized to 256 pixels. 
Then, a $256 \times 256$ center crop is taken, followed by a random horizontal flip, followed by embedding with the Stable Diffusion autoencoder. A similar process is done for the $512^2$ model. At test time, we follow the same preprocessing pipeline, but omit the random horizontal flip. Classification performance could improve if stronger augmentations, like RandomResizedCrop or color jitter, are used during the diffusion model training process.

\subsection{Baselines for Zero-Shot Classification}
\label{sec:baseline_details}

\paragraph{Synthetic SD Data:} We provide the implementation details of the ``Synthetic SD Data'' baseline (row 1 of Table~\ref{tab:zero_shot_cls}) for the task of zero-shot image classification.
Our Diffusion Classifier approach builds on the intuition that a model capable of generating examples of desired classes should be able to directly discriminate between them.
In contrast, this baseline takes the simple approach of using our generative model, Stable Diffusion, as intended to generate \emph{synthetic training data} for a discriminative model.
For a given dataset, we use pre-trained Stable Diffusion 2.0 with default settings to generate $10{,}000$ synthetic $512 \times 512$ pixel images per class as follows:
we use the English class name and randomly sample a template from those provided by the CLIP~\cite{radford2021learning} authors to form the prompt for each generation.
We then train a supervised ResNet-50 classifier using the synthetic data and the labels corresponding to the class name that was used to generate each image.
We use batch size $= 256$, weight decay $= 1e-4$, learning rate $= 0.1$ with a cosine schedule, the AdamW optimizer, and use random resized crop \& horizontal flip transforms.
We create a validation set using the synthetic data by randomly selecting 10\% of the images for each class; we use this for early stopping to prevent over-fitting. Finally, we report the accuracy on the target dataset's proper test set.

\paragraph{SD Features:}
We provide the implementation details of the ``SD Features'' baseline (row 2 of Table~\ref{tab:zero_shot_cls}) for the task of image classification. This baseline is inspired by Label-DDPM \cite{baranchuk2022labelefficient}, a recent work on diffusion-based semantic segmentation. Unlike Label-DDPM, which leverages a category-specific diffusion model, we directly build on top of the open-sourced Stable Diffusion model (trained on the LAION dataset). We then approach the task of classification as follows: given the pre-trained Stable Diffusion model, we extract the intermediate U-Net features corresponding to the input image. These features are then passed through a ResNet-based classifier to predict logits for the potential classes.
To extract the intermediate U-Net features, we add a noise equivalent to the $100th$ timestep noise to the input image and evaluate the corresponding noisy latent using the forward diffusion process. We then pass the noisy latent through the U-Net model, conditioned on timestep $t = 100$ and text conditioning ($\mathbf{c}$) as an empty string, and extract the features from the mid-layer of the U-Net at a resolution of [8 × 8 × 1024]. Next, we train a supervised classifier on top of these features. \textit{Thus, this baseline is not zero-shot.} The architecture of our classifier is similar to ResNet-18, with small modifications to make it compatible with an input size of  $[8 \times 8 \times 1024]$. Table \ref{table:resnet18sd} defines these modifications. We set batch size $= 16$, learning rate $= 1e-4$, and use the AdamW optimizer. During training, we apply image augmentations typically used by discriminative classifiers (RandomResizedCrop and horizontal flip). We do early stopping using the validation set to prevent overfitting.

\begin{table}
\centering
\begin{tabular}{c c c c c c} 
 \toprule
 Arch & Conv1 & Conv2 & Conv3 x2 & Conv4 x2 & Conv5 x2 \\ [0.5ex] 
 \hline
 ResNet-18 & 7x7x64 & 3x3 max-pool & 3x3x128 & 3x3x256 & 3x3x512 \\ 
 ResNet-18 (SD Features) & 3x3x1280 & - & 3x3x1280 & 3x3x2560 & 3x3x2560 \\
 \bottomrule
\end{tabular}
\vspace{0.3cm}
\caption{Comparison of SD Features' ResNet-18 classifier architecture with the original ResNet-18}
\label{table:resnet18sd}
\vspace{-1em}
\end{table}

\section{Techniques that did not help}
\model requires many samples to accurately estimate the ELBO. In addition to using the techniques in Section~\ref{sec:method} and \ref{sec:practical}, we tried several other options for variance reduction. None of the following methods worked, however. We list negative results here for completeness, so others do not have to retry them. 

\begin{figure}[h!]
\centering
\begin{minipage}{0.49\textwidth}
  \centering
  \includegraphics[width=\linewidth]{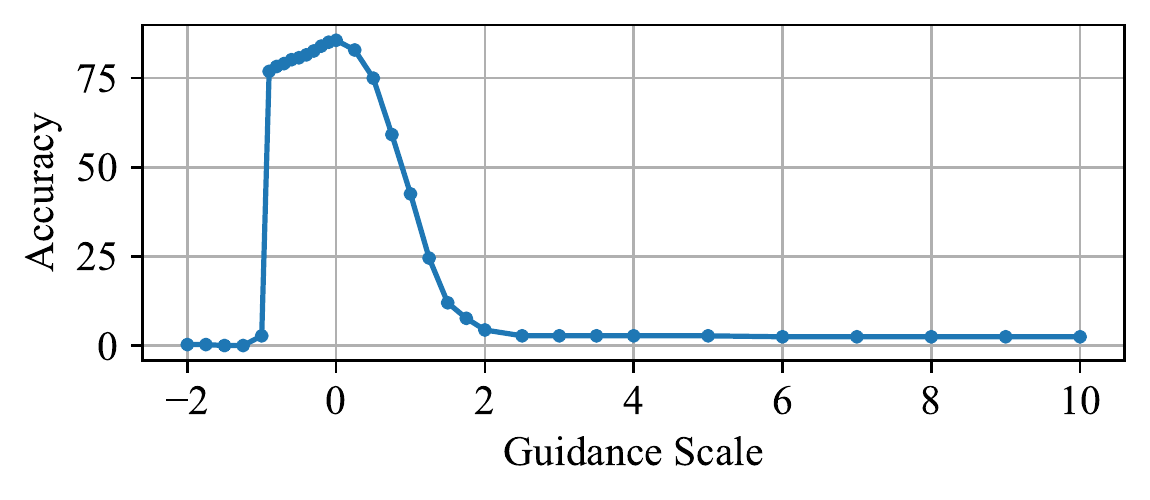}
  \caption{Accuracy plot of classifier-free guidance on Pets.}
  \label{fig:classifier_free}
\end{minipage}\hfill
\begin{minipage}{0.49\textwidth}
  \centering
  \includegraphics[width=\linewidth]{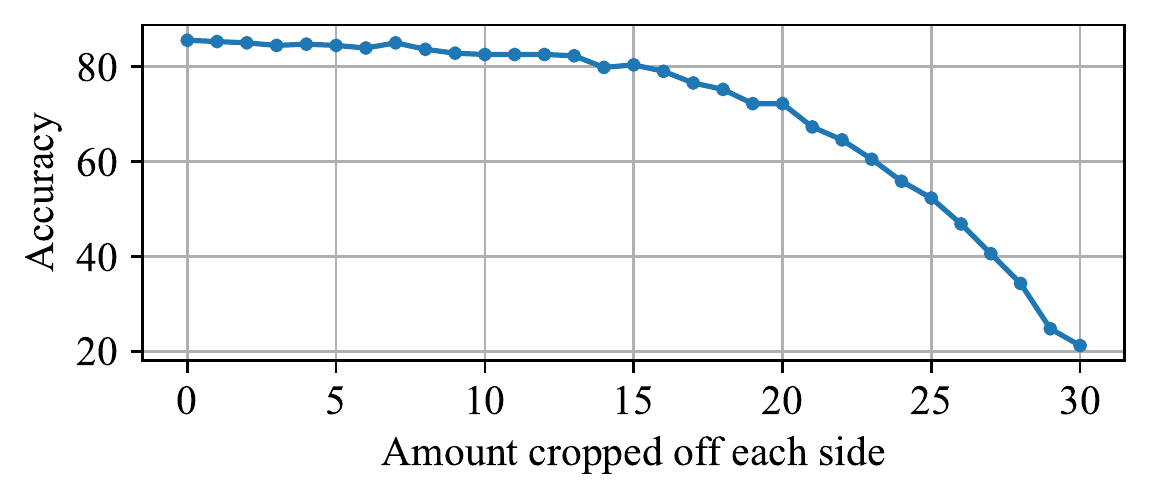}
  \caption{Cropping $\epsilon$ and $\epsilon_\theta(\mathbf{x}_t, \mathbf{c})$ reduces accuracy on Pets. }
  \label{fig:error_cropping}
\end{minipage}
\end{figure}

\paragraph{Classifier-free guidance}
Classifier-free guidance~\cite{ho2022classifier} is a technique that improves the match between a prompt and generated image, at the cost of mode coverage. This is done by training a conditional $\epsilon_\theta(\mathbf{x}_t, \mathbf{c})$ and unconditional $\epsilon_\theta(\mathbf{x}_t)$ denoising network and combining their predictions at sampling time: 
\begin{align}
    \Tilde{\epsilon}(\mathbf{x}_t, \mathbf{c}) = (1 + w) \epsilon_\theta(\mathbf{x}_t, \mathbf{c}) - w \epsilon_\theta(\mathbf{x}_t)
\end{align}
where $w$ is a guidance weight that is typically in the range $[0, 10]$. 
Most diffusion models are trained to enable this trick by occasionally replacing the conditioning $\mathbf{c}$ with an empty token. Intuitively, classifier-free guidance ``sharpens'' $\log p_\theta(x \mid  \mathbf{c})$ by encouraging the model to move away from regions that unconditionally have high probability. 
We test \model to see if using the $\Tilde{\epsilon}$ from classifier-free guidance can improve confidence and classification accuracy. Our new $\epsilon$-prediction metric is now 
    $\left\|\epsilon - (1 + w) \epsilon_\theta(\mathbf{x}_t, \mathbf{c}) - w \epsilon_\theta(\mathbf{x}_t) \right\|^2$.
However, \cref{fig:classifier_free} shows that $w=0$ (i.e., no classifier-free guidance) performs best. We hypothesize that this is because \model fails on uncertain examples, which classifier-free guidance affects unpredictably.

\paragraph{Error map cropping}

The ELBO $\mathbb{E}_{t, \epsilon}[\|\epsilon - \epsilon_\theta(\mathbf{x}_t, \mathbf{c})\|^2]$ depends on accurately estimating the added noise at every location of the $64 \times 64 \times 4$ image latent. We try to reduce the impact of edge pixels (which are less likely to contain the subject) by computing $\mathbf{x}_t$ as normal, but only measuring the ELBO on a center crop of $\epsilon$ and $\epsilon_\theta(\mathbf{x}_t, \mathbf{c})$. We compute: 
\begin{align}
    \|\epsilon_{[i:-i, i:-i]} - \epsilon_\theta(\mathbf{x}_t, \mathbf{c})_{[i:-i, i:-i]}\|^2
\end{align}
where $i$ is the number of latent ``pixels'' to remove from each edge. However, Figure~\ref{fig:error_cropping} shows that any amount of cropping reduces accuracy. 

\paragraph{Importance sampling}
Importance sampling is a common method for reducing the variance of a Monte Carlo estimate. Instead of sampling $\epsilon \sim \mathcal{N}(0, I)$, we sample $\epsilon$ from a narrower distribution. We first tried fixing $\epsilon = 0$, which is the mode of $\mathcal{N}(0, I)$, and only varying the timestep $t$. We also tried the truncation trick~\cite{brock2018large}
which samples $\epsilon \sim \mathcal{N}(0, I)$ but continually resamples elements that fall outside the interval $[a, b]$. Finally, we tried sampling $\epsilon \sim \mathcal{N}(0, I)$ and rescaling them to the expected norm ($\epsilon \rightarrow \frac{\epsilon}{\|\epsilon\|_2}\mathbb{E}_{\epsilon'}[\|\epsilon'\|_2])$) so that there are no outliers. 
Table~\ref{tab:importance_sampling} shows that none of these importance sampling strategies improve accuracy. This is likely because the noise $\epsilon$ sampled with these strategies 
are completely out-of-distribution for the noise prediction model. For computational reasons, we performed this experiment on a 10\% subset of Pets.
\begin{table}[h!]
    \centering
    \begin{tabular}{lc}
        \toprule
        Sampling distribution for $\epsilon$ & Pets accuracy \\
        \midrule
        $\epsilon = 0$ & 41.3 \\
        TruncatedNormal, $[-1, 1]$ & 49.9 \\
        TruncatedNormal, $[-2.5, 2.5]$ & 81.5\\
        Expected norm & 86.9 \\
        $\epsilon \sim \mathcal{N}(0, I)$ & 87.5 \\
        \bottomrule
    \end{tabular}
    \vspace{2mm}
    \caption{Every importance sampling strategy underperforms sampling the noise $\epsilon$ from a standard normal distribution. }
    \label{tab:importance_sampling}
\end{table}

\end{document}